\documentclass{article} 
\usepackage{iclr2023_conference,times}

\usepackage{amsmath,amsfonts,bm}

\def\eqref#1{equation~\ref{#1}}

\def\1{\bm{1}}

\def\vh{{\bm{h}}}

\def\vr{{\bm{r}}}
\def\vs{{\bm{s}}}

\DeclareMathAlphabet{\mathsfit}{\encodingdefault}{\sfdefault}{m}{sl}
\SetMathAlphabet{\mathsfit}{bold}{\encodingdefault}{\sfdefault}{bx}{n}

\definecolor{mydarkblue}{rgb}{0,0.08,0.45}
\usepackage[colorlinks=true,linkcolor=mydarkblue,citecolor=mydarkblue,filecolor=mydarkblue,urlcolor=mydarkblue]{hyperref}

\usepackage{hyperref}
\usepackage{url}
\usepackage{color,soul}
\usepackage{xcolor}
\usepackage{makecell}

\usepackage[utf8]{inputenc} 
\usepackage[T1]{fontenc}    
\usepackage{nicefrac}       
\usepackage{microtype}      
\usepackage{xcolor}         
\usepackage{colortbl}

\usepackage{multirow} 
\usepackage{times}
\usepackage{latexsym}
\usepackage{makecell}
\usepackage{wrapfig}
\usepackage{microtype}
\usepackage{graphicx}
\usepackage{times}
\usepackage{latexsym}
\usepackage{amsmath}
\usepackage{dashrule}

\usepackage{bbm}
\usepackage{booktabs}
\usepackage{amsfonts}
\usepackage{amsmath}
\usepackage{amssymb}
\usepackage{mathtools}
\usepackage{nicefrac}
\usepackage{microtype}
\usepackage{enumitem}
\usepackage{subfigure}
\usepackage{floatrow}
\usepackage{extarrows}
\usepackage[tableposition=top]{caption}
\newfloatcommand{capbtabbox}{table}[][\FBwidth]
\usepackage{color}
\usepackage{xcolor}
\usepackage{wrapfig}
\usepackage{lipsum}
\usepackage{graphics}
\usepackage{algorithm}
\usepackage{algorithmic}
\usepackage{blindtext}
\usepackage{bchart}
\usepackage{longtable}

\definecolor{mydarkblue}{rgb}{0,0.08,0.45}
 \usepackage[colorlinks=true,linkcolor=mydarkblue,citecolor=mydarkblue,filecolor=mydarkblue,urlcolor=mydarkblue]{hyperref}
 
\usepackage{float}
\usepackage{footnote}
\usepackage{enumitem}
\usepackage{bm}
\usepackage{arydshln}
\usepackage{booktabs}
\usepackage{color}
\usepackage{cite}
\usepackage{geometry}
\usepackage{microtype}
\usepackage{amsfonts}
\usepackage{url}
\usepackage{bbding}
\usepackage{tabu}
\usepackage{tikz}
\usepackage{vcell}

\usepackage{hyperref}
\usepackage{pifont}
\newcommand{\cmark}{\ding{51}}%
\newcommand{\xmark}{\ding{55}}%

\newcommand{\hlc}[2][yellow]{{%
    \colorlet{foo}{#1}%
    \sethlcolor{foo}\hl{#2}}%
}

\newenvironment{itemize*}%
 {\leftmargini=10pt\begin{itemize}%
  \setlength{\itemsep}{0pt}%
  \setlength{\parskip}{0pt}%
  }%
 {\end{itemize}}
\newenvironment{enumerate*}%
 {\begin{enumerate}%
  \setlength{\itemsep}{0pt}%
  \setlength{\parskip}{0pt}}%
 {\end{enumerate}}

\newcommand{\PPL}{{\mathrm{PPL}}}
\definecolor{myblue}{rgb}{0.82, 0.94, 0.75}
\definecolor{myyellow}{rgb}{1.0, 1.0, 0.73}
\definecolor{mygreen}{rgb}{0.73, 1.0, 1.0}
\definecolor{mypink}{rgb}{1.0, 0.85, 0.88}
\def\arrvline{\hfil\kern\arraycolsep\vline\kern-\arraycolsep\hfilneg}
\definecolor{negative}{HTML}{CC0000}
\definecolor{positive}{HTML}{0000CC}
\definecolor{asli}{HTML}{246FFF}
\definecolor{maryam}{HTML}{12b637}
\definecolor{moya}{HTML}{FF3800}
\definecolor{olga}{HTML}{F0F8FF}
\definecolor{spencer}{HTML}{0038FF}
\definecolor{luke}{HTML}{FF5733}

\newcommand{\asli}[1]{{\textcolor{asli}{[asli: #1]}}}

\newcommand{\olga}[1]{{\textcolor{purple}{[olga: #1]}}}

\newcommand{\ourmodel}{\texttt{ROSCOE}}
\newcommand{\ourmodelsa}{\texttt{ROSCOE-SA}}
\newcommand{\ourmodelss}{\texttt{ROSCOE-SS}}
\newcommand{\ourmodelli}{\texttt{ROSCOE-LI}}
\newcommand{\ourmodellc}{\texttt{ROSCOE-LC}}

\newcommand{\bartscore}{BARTScore}
\newcommand{\bartscoreParaCnn}{BARTScore+}
\newcommand{\bartscoreFineTuned}{BARTScore-P}

\title{\ourmodel: A Suite of Metrics for Scoring Step-by-Step Reasoning}

\author{Olga Golovneva, Moya Chen, Spencer Poff, Martin Corredor, Luke Zettlemoyer, \\
\textbf{Maryam Fazel-Zarandi, Asli Celikyilmaz} \\
Meta AI Research \\
\texttt{\{olggol, mpchen, spoff, mcorredor, lsz, maryamfazel, aslic\}@meta.com }
}

\iclrfinalcopy 

\begin{document}

\maketitle

\begin{abstract}
Large language models show improved downstream task performance when prompted to generate \textit{step-by-step} reasoning to justify their final answers~\citep{nye2021show,wei2022chain}.
These reasoning steps greatly improve model interpretability and verification, but objectively studying their correctness (independent of the final answer) is difficult without reliable methods for automatic evaluation. We simply do not know how often the stated reasoning steps actually support the final end task predictions. 
In this work, we present \ourmodel, a suite of interpretable, unsupervised automatic scores that improve and extend previous text generation evaluation metrics.
To evaluate \ourmodel~against baseline metrics, we design a typology of reasoning errors and collect synthetic and human evaluation scores on commonly used reasoning datasets.
In contrast with existing metrics, \ourmodel~can measure semantic consistency, logicality, informativeness, fluency, and factuality — among other traits — by leveraging properties of step-by-step rationales. We empirically verify the strength of our metrics on five human annotated and six programmatically perturbed diagnostics datasets - covering a diverse set of tasks that require reasoning skills and show that \ourmodel~can consistently outperform baseline metrics.
\footnote{Code can be found at \url{https://github.com/facebookresearch/ParlAI/tree/main/projects/roscoe}. Annotated datasets can be downloaded from \url{https://dl.fbaipublicfiles.com/parlai/projects/roscoe/annotations.zip}.
}

\end{abstract}

\section{Introduction}
\label{intro}
Scaling language models has improved state-of-the-art performance on nearly every NLP benchmark \citep{lmasfewshotlearners},
with large language models (LLMs) performing impressively as few-shot learners~\citep{lmasfewshotlearners}. 
Despite these achievements, even the largest of these models still struggle with tasks including math word problems \citep{hendrycksmath2021}, symbolic manipulation~\citep{rytting2021leveraging}, and commonsense reasoning~\citep{west-etal-2022-symbolic}. Recent work has shown that prompting~\citep{wei2022chain, wang2022self} or fine-tuning~\citep{pmlr-v162-lampinen22a} LLMs to generate \textit{step-by-step} rationales can lead to improvements on reasoning tasks.  
Some of these include small-scale analysis of specific error types within step-by-step rationales~\citep{lewkowycz2022solving,chowdhery2022palm}, as shown in Table \ref{introfigure}. 
However, existing works primarily focus on end-task performance. 
Although text generation evaluation metrics sometimes offer fine-grained quality evaluations (e.g., adequacy, fluency) against human scores~\citep{opitz-frank-2021-towards,expmetric}, these metrics generally treat the output as a whole, and many of these generative metrics operate on tasks such as summarization or machine-translation rather than reasoning.

\setlength\intextsep{-0.4pt}
\begin{wraptable}[18]{r}{7cm}
\caption{\footnotesize Step-by-step reasoning for logical inference task generated from 2-shot GPT-3 (text-davinci-002) \citep{lmasfewshotlearners}. Error types are highlighted: \hlc[mypink]{logical or factual issues}, \hlc[myyellow]{partially correct}, \hlc[mygreen]{somewhat-relevant}. Full 2-shot exemplars are in Appendix~\ref{app:intro}.}
\small
\vspace{-10pt}
\fbox{
\scalebox{0.85}
	    {
\begin{minipage}{\textwidth}
\textbf{Context}: Shelley is from Virginia, but is visiting that city with that famous market where they throw the fish. Going home next Tuesday! \\
\textbf{Question}: Is it likely that Shelly will be near the Pacific Ocean this weekend?\\
\textbf{GPT-3 Step-by-step reasoning:}

\noindent $\ \ \ \bullet$ \hlc[mypink]{No, it is not likely that Shelley will be near the Pacific Ocean this weekend.} 

\noindent $\ \ \ \bullet$ \hlc[mygreen]{Shelley is from Virginia, which is on the East Coast of the United States.} 

\noindent $\ \ \ \bullet$ {The Pacific Ocean is on the West Coast.} 

\noindent $\ \ \ \bullet$ \hlc[myyellow]{Shelley is visiting a city with a famous market, which could be anywhere in the world.}

\noindent $\ \ \ \bullet$ \hlc[mypink]{Tuesday is her travel day, so she will likely not be traveling this weekend.}
\end{minipage}
}
}
\vspace{-5pt}
\label{introfigure}
\end{wraptable}

In this paper, we present ~\ourmodel, a suite of interpretable and fine-grained step-by-step generation evaluation metrics to address the above gaps.
Rather than providing one score that only evaluates the generated text on the overall, 
\ourmodel~encapsulates fine-grained metrics under four perspectives: (1) \textit{semantic alignment} defines to 
what extend the generated reasoning is coherent, and grounded with the source context; (2) \textit{logical inference} evaluates if the generated reasoning steps are consistent within itself and checks for logical fallacies; (3) \textit{semantic similarity} quantifies the degree of similarity between the generated reasoning and the context or between intermediate steps to capture hallucinations or repetitions; and (4) \textit{language coherence} evaluates if the whole chain flows naturally. 

To evaluate \ourmodel~against existing metrics, 
we devise a taxonomy of reasoning errors for multi-step generations and use it to create synthetic data and collect human evaluations on commonly used reasoning datasets.
Our taxonomy and annotated datasets help us gain deeper insights into the causes of
reasoning inconsistencies and weaknesses of LLMs.
We evaluate \ourmodel~with $18$ fine-grained metrics under the above four perspectives. \ourmodel~demonstrates performance gains against baseline evaluation metrics on all  
tasks that require reasoning over context. 
Additional sensitivity analysis shows that \ourmodel~ is more robust when dealing with tasks that require logical and arithmetic reasoning. 

\textbf{Contributions}. (1) We propose a new taxonomy for reasoning errors, and use it for collecting human annotations and creating synthetic datasets. (2) Using our taxonomy, we propose a new suite of metrics that focus on sequence and step level analysis of step-by-step reasoning. (3) We present extensive comparative analysis on 11 datasets of varied complex reasoning problems demonstrating the strengths of each metric, especially in terms of interpretability relative to baselines, and considerations for use. 

\section{Related Work}
\label{related}
\textbf{Evaluating Explanations.} 
Free-form natural Language (NL) explanations of model decisions should enable accurate representation of the reasoning process and degree of plausibility ~\citep{danilevsky-etal-2020-survey,jacovi-goldberg-2021-aligning,jacovi-etal-2021-contrastive}.
A qualitative assessment of NL explanations with correctness labels collected from human judges was presented in \citep{esnli}. 
Recent work has also investigated automatic metrics for natural language generation (NLG) evaluation including word overlap or embedding based similarly with  human written explanations~\citep{clinciu-etal-2021-study}. 
Though fast and cost-effective, automatic metrics for NLG are not equipped to measure the logical inconsistencies or information gain with thinking steps~\citep{reiter-2019-natural,surveyevaluation}. 
Explanations have also been evaluated by collecting datasets, 
and running correlation analysis to investigate
the degree to which an automatic
metric correlates with human judgements of clarity, relevance and informativeness~\citep{expmetric,welleck2022naturalprover}.
Although reliable, human evaluation is an expensive, domain specific, and time-consuming process. In comparison, \ourmodel~provides generic automatic evaluation
procedures that are domain and task specific.

\textbf{Automatic Metrics.} Many NLG evaluation metrics exist in the literature including ones based on: \textit{n}-gram match \citep{lin-2004-rouge}, regression \citep{sellam-etal-2020-bleurt},
embedding proximity \citep{BERTScore}, paraphrasing \citep{prism}, generation as an evaluator \citep{bartscore}; information alignment \citep{ctc}; among others. Although these metrics are easy to use, they evaluate the alignment of two texts as a whole and are not designed to assess individual reasoning steps. 
The closest metrics to ours are CTC \citep{ctc} and BARTScore \citep{bartscore}, as both introduce a set of interpretable metrics to evaluate the similarity between two texts. However,
\ourmodel~is unique in providing fine-grained interpretations of reasoning steps, determining contradictions, and identifying ordering issues in the reasoning narrative.   

\textbf{Self-Consistency with LLMs.} Recent work on improving LLMs performance on complex reasoning tasks uses an ensemble strategy called self-consistency~\citep{wang2022self}. This method samples a diverse set of reasoning paths from a language model via reasoning traces prompting and returns the most consistent final answer in the set. Other work evaluates the diversity of a reasoning path~\citep{reasoner}, or the consistency of an inference step~\citep{selectioninference} or finetune LLMs~\citep{star} to improve on difficult NLP tasks. In contrast to these works, we present a suit of metrics that focus on determining
the type of the error (e.g., commonsense or logical inconsistency) in a reasoning path, if one exists. 

\section{Reasoning Error Taxonomy and Datasets Construction}
\label{taxonomy}
\textbf{Problem Formulation}.
\label{formulation}
Our goal is to score step-by-step rationales generated by a language model. We assume that the model is given a \textit{source} context $\vs = \{s_1, \cdots, s_T\}$ of \textit{T}-sentences indicating a problem statement followed by a question and is prompted to generate step-by-step reasoning \citep{nye2021show}. We refer to this as a \textit{hypothesis} $\vh = \{h_1, \cdots, h_{N}\}$ of \textit{N}-steps, including a final answer as the last step. 
We do not assume availability of gold step-by-step reasoning \textit{references} $\vr = \{r_1, \cdots, r_{K}\}$ of \textit{K}-steps.

\textbf{Taxonomy}. 
We propose a new taxonomy of generic reasoning errors for language problem solving. 
We first conduct manual preliminary analysis on different types of LLMs reasoning errors using five \textit{Human judged} datasets described below. 
Based on our analysis, we identified nine error types centered on the overall reasoning chain (i.e., the quality of the step-by-step thinking, including consistency with the context and commonsense reasoning).
Our taxonomy also includes fine-grained errors marking inconsistency of a reasoning step with the previous steps, whether each step contributes to the final decision, 
and overall logical inference or fluency issues. The definition of error types is in Table~\ref{tab:taxonomy-simple}, and Table~\ref{tab:apptaxonomy} provides examples. 

\begin{table}[t]
 \vspace{-5pt}
    \centering
    \scalebox{0.75}
	    {
    \begin{tabular}{lp{480pt}}
        \toprule
         \textbf{Error Type} & \textbf{Definition}\\
         \midrule
         \textbf{Grammar} 
& Faulty, unconventional, or controversial grammar usage \\
\textbf{Factuality}
& Information about an object (i.e. quantity, characteristics) or a named entity doesn't match with the input context. \\

\textbf{Hallucination}
&  Information is not provided in the problem statement and is irrelevant or wrong \\

\textbf{Redundancy}
&  Explanation contains redundant information, which even though might be factual, is not required to answer the question \\
\textbf{Repetition}
& Step paraphrases information already mentioned in previous reasoning steps \\

\textbf{Missing step}
&  The content of the generated reasoning is incomplete and lacks required information to produce the correct answer. \\

\textbf{Coherency}
& Steps contradict each other or do not follow a cohesive story \\

\textbf{Commonsense}
&  Model lacks relations that should be known from general world (e.g., "all ducks are birds") \\

\textbf{Arithmetic}
&  Error in math calculations \\
         \bottomrule
    \end{tabular}
    }
    \caption{\footnotesize Taxonomy of Step-by-Step Reasoning Errors. Full list of the error types with examples is illustrated in Table~\ref{tab:apptaxonomy}.} 
    \label{tab:taxonomy-simple}
\end{table}

\textbf{Datasets and Annotations}. To evaluate \ourmodel, we select datasets covering diverse set of tasks that require reasoning skills (e.g., logical, arithmetic, and commonsense reasoning tasks).
We separate these datasets into two: (1) \textbf{Diagnostics} datasets that contain gold standard step-wise reasoning chains, where we synthetically perturb some of the reasoning steps to introduce different generation errors (e.g., missing step, mathematical error, etc.); (2) \textbf{Human judged} datasets with model generated step-by-step reasoning outputs where the reasoning error evaluations are solicited from expert judges. We investigate these in $\S$\ref{experimentsetup}.
\section{Reasoning Scorer: \ourmodel}
\label{roscoethemetric}
We present our fine-grained metrics under four perspectives: \textit{semantic alignment}, \textit{semantic similarity}, \textit{logical inference} and \textit{language coherence}. Each metric is bounded within $[0, 1]$, where $1$ indicates the perfect score and $0$ corresponds to failure. A metric is \textit{reference-free} or \textit{unsupervised} when it uses the source and hypothesis ($\vh \rightarrow \vs$), while \textit{reference-based} or \textit{supervised} when evaluated between hypothesis and reference ($\vh \rightarrow \vr$).


\subsection{Semantic Alignment Metrics (\ourmodelsa)}
At the core of the \ourmodel~ semantic alignment\footnote{Semantic alignment refers to determination of relations between concepts with the same or a similar intended meaning~\citep{agirre-etal-2013-sem}.} metrics is the reasoning alignment vector from the $N$-step hypothesis $\vh$ to the source $\vs$ of length $T$:
$r\textnormal{-align}(\vh \rightarrow \vs) = \{\alpha_1, \alpha_2, \cdots, \alpha_N\}$,
where each alignment value $\alpha_i = r\textnormal{-align}(h_i \rightarrow \vs) = [1+\max_{j=1}^{T}(\cos(h_i, s_j)]/2 \in[0, 1]$ is the normalized cosine similarity between hypothesis step and most similar sentence in a context, and explicitly measures the grounding of the step-wise reasoning with respect to the source text (illustrated in App.~\ref{app:roscoe}, Fig.~\ref{fig:alignment}). 
We estimate the alignment vector $r\textnormal{-align}(\vh \rightarrow \vs)$ 
by matching source text and the reasoning chains on the embeddings of tokens and individual reasoning steps.
A similar information alignment score is introduced in CTC~\citep{ctc} to measure the confidence that the information of the $i$-th source document token $s_j$ is grounded by a hypothesis token $h_i$. Our reasoning alignment is different in that we measure if a hypothesized reasoning step $h_i$ supports the source context $\vs$. Our proposed metrics are summarized in Table~\ref{tab:sa-equations}.

\vspace{6pt}
\begin{table*}[h]
    \vspace{-10pt}
	\caption{\footnotesize Semantic alignment metrics (\ourmodelsa).}
	\label{tab:sa-equations}
	\begin{center}
	   \scalebox{0.7}
    {
     \begin{tabular}{>{\raggedright}p{3.5cm}p{17cm}}
    \toprule
         \textbf{Score} & \textbf{Description}\\
    \midrule
        \textbf{Faithfulness-Step} ($\vh \rightarrow \vs$) &
        This step-level score is based on the alignment from the hypothesis steps to the source sentences, and is calculated as the mean reasoning alignment score over the steps of reasoning (see illustration in Appendix~\ref{app:roscoe}, Figure~\ref{fig:alignment}): $(1/N)\textstyle\sum_{i=1}^{N}r\textnormal{-align}(h_i \rightarrow \vs)$. Faithfulness measures if the model misinterpreted the problem statement, or the reasoning chain is too vague, irrelevant, or misuses information.\\ 
    \midrule
        \textbf{Faithfulness-Token} ($\vh \rightarrow \vs$) & We extend step-level embeddings of the Faithfulness-Step by measuring similarities between the token embeddings: $(1/(N + M))\textstyle\sum_{i=1}^{N} [r\textnormal{-align}(h_i \rightarrow \vs)$ 
        $ + \textstyle\sum_{j=1}^{M_i}r\textnormal{-align}^{token}(h_{i,j} \rightarrow \vs)]$,
 as shown in App.~\ref{app:roscoe}, Fig.~\ref{fig:alignment}.
 $M_i$ is the number of tokens in step $h_i$, $M = \sum_{i=1}^{N} M_i$ is the total number of tokens in the reasoning chain, $h_{i,j}$ is the \textit{j}th token in \textit{i}th step, and $r\textnormal{-align}^\textit{token}$ is the alignment vector from tokens in step $h_{i}$ to all tokens in $\vs$.\\
    \midrule
         \textbf{Informativeness-Step (Info-Step)} ($\vh \leftrightarrow \vs$) & Measures how well information present in the source is used in the reasoning steps: $[(1/T)\textstyle\sum_{t=1}^{T}r\textnormal{-align}(s_t \rightarrow \vh) + (1/N)\textstyle\sum_{i=1}^{N}r\textnormal{-align}(h_i \rightarrow \vs)] / 2$. Info-step gives a higher score to reasoning steps that are well-grounded with respect to the source, and identifies the degree of information from source that is covered by the generated hypothesis. A lower Info-Step score corresponds to the reasoning steps that are not related to the source sentences or have missed information provided in the context. \\
    \midrule
        \textbf{Repetition-Token} ($h_i \rightarrow h_j$) & To identify repeated, or paraphrased steps, we look at the token alignment scores between all steps in the hypothesis chain: $1 - \textstyle\max_{i=2..N} \textstyle\max_{j=1\cdots i-1}[  (1/M_i)\textstyle\sum_{l=1}^{M_i} r\textnormal{-align}^{token}(h_{i,l} \rightarrow h_j) ]$. For each pair of sentences, we look at the mean token alignment, and find those sentences that maximize this alignment score. In other words, Repetition-Token will punish chains where there are at least two steps with high overlap in token embeddings. \\
    \midrule
        \textbf{Hallucination} ($\vh \rightarrow (\vs,\vr)$)
        & To find irrelevant reasoning steps, we use alignment score to identify steps that are both not related to the context and not in the reference chain (to avoid punishing for possibly relevant commonsense knowledge): $1 - \textstyle\max_{i=1..N}([\mathbf{1} - r\textnormal{-align}(\vh \rightarrow \vs)]\cdot[\mathbf{1} - r\textnormal{-align}(\vh \rightarrow \vr)])$. Here, $\mathbf{1}$ is an all-ones vector, and ($\cdot$) is the element-wise product. \\
    \midrule
        \textbf{Redundancy} ($\vh\rightarrow\vr$)
        & To find chains that contain information that is not required to solve the problem (i.e., redundant steps), we identify those hypothesis steps that are least aligned with the the reference steps: $\textstyle\min_{i=1..N} r\textnormal{-align}(h_i \rightarrow \vr)$. This score punishes chains with steps that are not required for the correct solution. \\
    \midrule
        \textbf{Semantic Coverage-Step} ($(\vr, \vh) \rightarrow \vs$)
        & This score can be viewed as a measure of how easily a gold reference could be generated by the hypothesis. It compares step level grounding of the hypothesis with respect to the source, and the gold reference grounding:
$\lvert
 (1/T)\textstyle\sum_{t=1}^{K}r\textnormal{-align}(r_t\rightarrow \vs) - (1/N)\textstyle\sum_{i=1}^{N}r\textnormal{-align}(h_i \rightarrow \vs)
 \rvert$,
 where $\lvert \cdot \rvert$ indicates absolute value. \\ 
    \midrule
        \textbf{Reasoning Alignment} ($\vh \rightarrow \vr$)
        & The most straightforward way to evaluate the correctness of the hypothesis chain is to compare the degree of the overlap between the hypothesis and the reference. One way of doing that is to measure the reasoning alignment between them: $(1/N)\textstyle\sum_{i=1}^{N}r\textnormal{-align}(h_i \rightarrow \vr)$. \\ 
    \midrule
        \textbf{Commonsense} \hspace{10pt}($\vr \rightarrow (\vh, \vs)$) & Measures if hypothesis lacks steps that are not stated in the source, but are required to solve the problem such as general world knowledge (e.g., "velocity is distance divided by time", "1 foot is 12 inches", "all ducks are birds", etc.). We detect such information by extracting steps in the reference reasoning that are not grounded by the source text: $1 - \textstyle\max_{i=1..K}( 
 [\mathbf{1} - r\textnormal{-align}(\vr \rightarrow \vh)]
 \cdot
 [\mathbf{1} - r\textnormal{-align}(\vr \rightarrow \vs)]
 )$. \\
    \midrule
        \textbf{Missing Step} ($\vr$$\rightarrow$$\vh$)
        & To identify steps that are missing from the hypothesis but could be required to solve the problem, we look at the alignment between reference and the hypothesis, similar to \textit{Redundancy}. However, here we go through each step in the reference, and check if there is a similar step in the hypothesis: $\textstyle\min_{i=1..K}(r\textnormal{-align}(r_i$$\rightarrow$$\vh))$. \\ 
    \bottomrule
    \end{tabular}
    }
    \vspace{-8pt}
	\end{center}
\end{table*}

\subsection{Semantic Similarity Metrics (\ourmodelss)}
Semantic similarity metrics quantify the degree of semantic equivalence between pieces of text. As opposed to the \ourmodelsa~metrics, \ourmodelss~considers text as a whole, rather than relying on text units comparisons. We propose the following metrics summarized in Table~\ref{tab:ss-equations}.

\vspace{10pt}
\begin{table*}[h]
    \vspace{-10pt}
	\caption{\footnotesize Semantic similarity metrics (\ourmodelss).}
	\label{tab:ss-equations}
	\begin{center}
	   \scalebox{0.71}
    {
     \begin{tabular}{>{\raggedright}p{3.5cm}p{17cm}}
    \toprule
         \textbf{Score} & \textbf{Description}\\
    \midrule
        \textbf{Informativeness-Chain (Info-Chain)} ($\vh$$\rightarrow$$\vs$) &
        Similar to Info-Step, this metric quantifies the degree of agreement between the hypothesis chain and the source and is calculated as $[1 + \cos(\vh, \vs)] / 2$. We embed reasoning chain and source context as a whole, as opposed to using step-wise embeddings in *-Step types of metrics introduced in Table~\ref{tab:sa-equations}. 
        \\
    \midrule
        \textbf{Repetition-Step} ($h_i$$\leftrightarrow$$h_j$) &
        Measures repetition-related errors on the step level by checking if it paraphrases information already mentioned in the previous steps: $(1 - \textstyle\max_{i=2..N} \textstyle\max_{j=1\cdots i-1} [\cos(h_i, h_j)]) / 2$. Unlike Repetition-Token, which is orderless and compares individual tokens in pairs of steps, Repetition-Step considers step embeddings similarity and is more robust to changing contexts. \\
    \midrule
        \textbf{Semantic Coverage- Chain} ($\vr \leftrightarrow \vh$) & 
        Reflects the overall degree of similarity between the reference and hypothesis chains, comparing reference and hypothesis embeddings as a whole: $[1 + \cos(\vr, \vh)] / 2$.
        \\
    \bottomrule
    \end{tabular}
    }
	\end{center}
\end{table*} 

\subsection{Logical Inference Metrics (\ourmodelli)}

Logical inference metrics (Table~\ref{tab:li-equations}) measure logical errors between pieces of text. We use an NLI model that was trained to classify hypothesis-context pairs into entailment, neutral, and contradiction classes~\citep{laurerless} to infer the contradiction probability $p_{\mathrm{contr}}$.

\vspace{10pt}
\begin{table*}[h]
    \vspace{-10pt}
	\caption{\footnotesize Logical inference metrics (\ourmodelli).}
	\label{tab:li-equations}
	\begin{center}
	   \scalebox{0.71}
    {
     \begin{tabular}{p{3.5cm}p{17cm}}
    \toprule
         \textbf{Score} & \textbf{Description}\\
    \midrule
        \textbf{Self-Consistency} ($h_i$$\leftrightarrow$$h_j$) &
        Measures logical entailment errors \textit{within} the reasoning steps: $1-\textstyle\max_{i=2..N} \textstyle\max_{j<i}p_{\mathrm{contr}}(h_i , h_j)$. This metric will punish chains where there is a pair of steps that are likely to contradict each other. \\
    \midrule
        \textbf{Source-Consistency} ($\vh \leftrightarrow \vs$) & Measures logical entailment errors between any generated reasoning $\vh$ and the source context $\vs$: $1 - \textstyle\max_{i=1..N} \textstyle\max_{j=1..T} ~p_{\mathrm{contr}}(h_i , s_j)$. Specifically, for each reasoning step we measure the probability that it contradicts any sentence in the context. We take the maximum probability of contradiction over all steps, following the logic that a contradiction anywhere in the reasoning chain signals a failure of the overall argument. \\
    \bottomrule
    \end{tabular}
    }
	\end{center}
\end{table*}

\subsection{Language Coherence Metrics (\ourmodellc)} 
To evaluate language coherence (Table~\ref{tab:lc-equations}), we use perplexity $\PPL$ as scored by the GPT2-Large model \citep{radford2019language}, and English grammatical acceptability $p_{\mathrm{gram}}$ as scored by the classifier model from \citet{krishna2020style}. Both models were used as-is with no finetuning.

\begin{table*}[h]
    \vspace{-10pt}
	\caption{\footnotesize Language coherence metrics (\ourmodellc).}
	\label{tab:lc-equations}
	\begin{center}
	   \scalebox{0.71}
    {
     \begin{tabular}{p{3.5cm}p{17cm}}
    \toprule
         \textbf{Score} & \textbf{Description}\\
    \midrule
        \textbf{Perplexity-Chain} ($\vh$) & Average perplexity of all tokens in the generated reasoning steps: $1 / \PPL(\vh)$. The context used to score each token is the previous tokens in the current and all previous steps. Steps are joined with a space character. To keep the range and orientation consistent with the other scores we invert the perplexity.
        \\
    \midrule
        \textbf{Perplexity-Step} ($h_{i}$) &
        Average perplexity of all tokens in the generated reasoning steps, where the context used to score each token is only the previous tokens within the current step: $1 / [ (1/N) \textstyle\sum_{i=0}^{N} \PPL(h_i) ]$. To keep the range and orientation consistent with the other scores we invert the perplexity.\\
    \midrule
        \textbf{Grammar} ($h_{i}$) &
        Probability of grammatical acceptability of each step, averaged over all steps: $(1/N) \textstyle\sum_{i=0}^{N} p_{\mathrm{gram}}(h_i)$. \\
    \bottomrule
    \end{tabular}
    }
	\end{center}
\end{table*} 

\section{Experimental Setup}
\label{experimentsetup}
\textbf{Diagnostics Datasets}. 
\label{datasets}
We construct our first category of labeled datasets by generating perturbations --- i.e., deterministic modifications --- on half of the reference reasoning steps and assign binary labels based on whether or not a chain has been perturbed. We select seven language understanding and entailment datasets that require complex problem solving skills, and have reference step-by-step explanations: \textbf{Entailment-Bank} (deductive reasoning)~\citep{entalmentbank2021}, \textbf{ProofWriter} (logical reasoning) \citep{tafjord-etal-2021-proofwriter}; three arithmetic reasoning datasets \textbf{MATH}~\citep{hendrycksmath2021}, \textbf{ASDIV}~\citep{miao-etal-2020-diverse} and \textbf{AQUA}~\citep{liang-etal-2018-meaning}; \textbf{EQASC} (explanations for commonsense question answering) \citep{aggarwaletal2021ecqa}, and \textbf{StrategyQA} (question answering with implicit reasoning strategies) \citep{geva2021did} (see dataset details in App.~\ref{app:diagnosticdatasets}). 
Using our taxonomy, we introduce 12 error perturbation rules and apply on these datasets to construct our diagnostics datasets (see details in App.~\ref{app:datasetconstruction}).

\textbf{Human Judged Datasets}. We select our second category of datasets from commonly used complex reasoning tasks: \textbf{GSM8K} (arithmetic reasoning)~\citep{cobbe2021gsm8k}, \textbf{DROP} (discrete reasoning)~\citep{dua-etal-2019-drop}, \textbf{ESNLI} (deductive and commonsense reasoning)~\citep{esnli}, \textbf{COSMOS-QA }(commonsense reasoning)~\citep{huang-etal-2019-cosmos} and \textbf{SemEVAL}~\citep{Ostermann2018SemEval2018T1} (commonsense reasoning). 
\citeauthor{wei2022chain} (\citeyear{wei2022chain}) provide model generated chain of thought reasoning steps for GSM8K. We used chains produced by the \textit{175b\_verification} model to annotate for reasoning errors.
For other datasets, we prompt GPT-3 LLM~\citep{lmasfewshotlearners}
with few-shot in-context examples to obtain step-by-step reasoning sequences (see examples in App.~\ref{app:humanjudgeddatasets}).
We use the error types in our taxonomy in Table~\ref{tab:taxonomy-simple} as human evaluation perspectives of reasoning errors where we solicit five expert annotators\footnote{We chose expert annotators over crowd-sourcing, because our annotation task is cognitively challenging and requires fine-grained annotation.}. 
The data collection interface provided judges with the source text (e.g., source and a question, or hypothesis, premise, and a question if they entail) and associated reasoning text clearly separated into individual steps. Judges were asked to rate the chain as a whole (e.g., on overall quality)
as well as each individual step (e.g., commonsense errors, contradicts with the previous steps). App. Table~\ref{tab:human-eval-error-stats} summarizes the distribution of error types annotated by the judges. See App.~\ref{app:annotation} for details.

\textbf{\ourmodel~Training.}
\label{training}
To obtain reasoning step embeddings, we finetune SimCSE~\citep{gao2021simcse}, a supervised sentence similarity model extending the RoBERTa word embedding model~\citep{liu2019roberta} on multi-step reasoning datasets we listed in~$\S$\ref{experimentsetup} (see details in Table~\ref{tab:dataset-appendix})\footnote{Fine-tuned model is available at \url{https://huggingface.co/facebook/roscoe-512-roberta-base}}. SimCSE is a contrastive learning model that is trained on triplets of reference reasoning steps, positive and hard-negative hypothesis reasoning steps to minimize the cross-entropy objective with in-batch negatives. For contrastive learning, we use the context and reference reasoning steps as a positive sample $(\vs, \vr)$, and context and perturbed reference steps $(\vs, \vh)$ as hard-negative pairs. For finetuning, we embed source context and hypothesis chain as a whole, without splitting it into steps. With the finetuned model we embed each individual step, as well as a reasoning chain as a whole. 
We use the pretrained checkpoint of supervised SimCSE model \textit{sup-simcse-roberta-base} to initialize our model, and further train it for five epochs on our synthetic train data
(details in App.~\ref{app:finetuning}). We also compare \ourmodel~scores calculated against \textit{sup-simcse-roberta-base} SimCSE model, and \textit{all-mpnet-base-v2} sentence embedding model~\citep{reimers-2019-sentence-bert} to understand metrics sensitivity to the embedding method.

\textbf{Baseline Metrics.} We use text generation evaluation metrics as baseline metrics and comprehensively examine the ones outlined in \S\ref{related}, which are: \textit{n}-gram match based metrics including \textbf{ROUGE-}1, \textbf{ROUGE-2}, and \textbf{ROUGE-L}~\citep{lin-2004-rouge}; pre-trained scores including \textbf{BLEURT}~\citep{sellam-etal-2020-bleurt}, \textbf{PRISM}~\citep{prism}, \textbf{BERTScore}~\citep{BERTScore}, \textbf{\bartscore} using the \textit{Faithfulness} ($\vs \rightarrow \vh$)  direction for factuality and relevance, and its finetuned variant 
BARTScore+CNN+Para \textbf{\bartscoreParaCnn} ~\citep{bartscore}; and information alignment metrics of \textbf{CTC}, 
\textbf{CTC-Relevancy} and \textbf{CTC-Consistency}. We also include \textbf{\bartscoreFineTuned}, which we obtain by finetuneing BART~\citep{lewis-etal-2020-bart} on the same reasoning datasets we use for finetuning our SimCSE embedding models. Most of our \ourmodel~metrics are constructed reference-free. We also have metrics that use reference reasoning steps which we examine against human judgements. We use the official code for each metric. 

\textbf{Meta Evaluation.} 
We use \textit{Somers'} $D$\footnote{We use SciPy \citep{2020SciPy-NMeth} to calculate correlations and obtain \textit{p-values} from a hypothesis test where the null hypothesis is an absence of association.}~\citep{somers1962new}, which measures the ordinal association between two measured quantities, to meta-evaluate each scorer against synthetic and human scores. We prefer \textit{Somers' $D$} over more commonly used \textit{Kendall’s $\tau$} or \textit{Kendall’s $\tau\textit{-}b$}, because  it is better in handling the ties of a biased random variable~\citep[Section 7.1.5]{A10}, which imposes an upper bound on the possible values \textit{Kendall’s $\tau(\textit{-}b)$} can take. For each score $Y$ considered, our correlations are built against the biased random variable $X\in[0,1]$, represented by the perturbation or error presence indicator and evaluated using $D(Y|X)=\tau(X,Y)/\tau(X,X)$.
\section{Experimental Results}
\label{experiments}
\textbf{Controlled Experiments with Diagnostics Datasets}.
\label{diagnostics-experiments}
Table~\ref{tab:corr-synth-unsup-simple} shows Somers’ $D$ correlation for metrics measured reference-free on six different datasets and compares baselines to \ourmodel-*~ aggregated categories calculated with finetuned embeddings: \ourmodelsa, \ourmodelss, \ourmodelli, \ourmodellc.
Results also include \ourmodel~metrics with \textit{all-mpnet-base-v2} (\ourmodelsa$^1$, \ourmodelss$^1$) and \textit{sup-simcse-roberta-base} (\ourmodelsa$^2$, \ourmodelss$^2$) sentence embedding models.
 Correlations for ProofWriter are taken on its \textit{depth-5} subset. We report highest correlation scores across perturbations within each dataset. 
The breakdown of all \ourmodel~metrics is in App. Table~\ref{tab:corr-synth-unsup-max}.

\begin{wraptable}{r}{8.5cm}
  \centering
  \footnotesize
  \setlength\tabcolsep{3.4pt}
    \vspace{-8pt}
  \renewcommand{\arraystretch}{1}
        \caption{\footnotesize Somers' $D$ correlation of different metrics on six \textbf{Diagnostics} datasets. Metrics are measured \textbf{reference-free} on $(\bm{s},\bm{h})$. We take the maximum score over different perturbations.  
        The two highest correlations for each dataset are \textbf{bolded} and \underline{underlined}, respectively. Correlations that are not significant (\textit{p-value} $\ge 0.05$) are omitted when aggregating, and "-" denotes an absence of any significant correlation. Breakdown of all baseline and \ourmodel~ metrics is shown in App.~\ref{app:diagnostics-experiments}, Table~\ref{tab:corr-synth-unsup-max}.
        }
    \scalebox{0.86}
    {
        \begin{tabular}{@{}l@{}c@{}c@{}c@{}c@{}c@{}c@{}}
    \toprule
          & \textbf{\hyperref[tab:corr-synth-unsup-perturb-entailment_bank]{EntBank}} \ \ &
          \textbf{\hyperref[tab:corr-synth-unsup-perturb-math_dataset]{Math}} \ \ & 
          \textbf{\hyperref[tab:corr-synth-unsup-perturb-aqua]{AQUA}} \ \ & 
          \textbf{\hyperref[tab:corr-synth-unsup-perturb-proofwriter]{ProofWr.}} \ \ & 
          \textbf{\hyperref[tab:corr-synth-unsup-perturb-eqasc]{EQASC}} \ \ & 
          \textbf{\hyperref[tab:corr-synth-unsup-perturb-asdiv]{ASDIV}} \\ 
          \midrule
    ROUGE-L  & 0.365 & 0.156 & 0.264 & 0.106 & 0.315 & 0.269\\
    BLEURT  & 0.257 & 0.148 & 0.252 & 0.024 & 0.447 & - \\
    BERTScore  & 0.380 & 0.124 & 0.220 & 0.117 & 0.462 & 0.322\\
    \bartscore  & 0.358 & 0.185 & 0.317 & 0.081 & 0.415 & - \\
    \bartscoreParaCnn  & 0.315 & 0.164 & 0.251 & 0.054 & 0.297 & - \\
    \bartscoreFineTuned  & 0.186 & 0.128 & 0.215 & 0.011 & 0.276 & - \\
    PRISM  & 0.453 & 0.208 & 0.191 & 0.235 & 0.436 & - \\
\bottomrule
    \end{tabular}
    }
  \label{tab:corr-synth-unsup-simple}%
\end{wraptable}%
We observe that: (1) \ourmodel~ can
outperform all other reference-free methods on all six diagnostic datasets, 
(2) the gains for \ourmodelss~ are more pronounced in four
out of six diagnostics datasets, which suggests that \ourmodel~can capture hallucinations and repetitions in step-wise reasoning. On Proofwriter, our scorers show lower correlations, because as shown in Table~\ref{app:diagnosticdatasets}, the context is a list of facts and rules and the reasoning steps can include unordered fact and rule combinations, but still a correct answer can be deduced. This makes it challenging for \ourmodel~ to evaluate the steps in sequence. 
Overall, the correlations of the baseline metrics are much lower than \ourmodel, because the baseline metrics are designed to capture the semantic or lexical overlap between a reference and hypothesis and it is harder to detect logical consistency without a golden reference text. \ourmodel~is specifically focused on reference-free settings, and can gauge each individual step against the source and other generated steps. In fact, our metrics also work well against the baselines in the reference-based setting (comparing against reference reasoning steps). In App. Table~\ref{tab:corr-synth-sup-max} we present correlations when 
metrics are measured as reference-based. 
We also observe that finetuning SimCSE gives highest improvements on the ASDIV dataset.
ASDIV is a 1-step reasoning dataset (see App. Table~\ref{alldatasetssamples-diaognostic}), where step is represented by an equation with one of the arithmetic perturbations added. We hypothesize that including these patterns in finetuning helped the model to better learn relationships between context and equations, and resulted in higher scores.
On EQASC dataset, Repetition* scores are able to catch all duplicated steps in a chain, i.e., we can separate perturbed and non-perturbed chains based on the given threshold value for the Repetition* scores, 
and achieve perfect correlation scores (App. Table~\ref{tab:corr-synth-unsup-perturb-eqasc}). 
To understand if finetuning actually helps to improve scoring, we compare non-aggregated metrics (see details in App. Table~\ref{tab:corr-synth-unsup-max}). We observe, that finetuning indeed helps to improve \ourmodel: on average across datasets, all correlations except Repetition\_* scores improve (up to $0.556$ on Informativeness-Chain), with mean Repetition-Token not changing, and mean Repetition-Step degrading by $0.005$. We speculate that since we finetune the model using reasoning chains and context as a whole, it helps to better capture step-by-step rationales, while possibly degrading on word and sentence-level semantics. 
{
\parfillskip=0pt
\parskip=0pt
\par}
\begin{wraptable}{r}{9cm}
    \vspace{-14pt}
	\caption{\footnotesize Somers' $D$ correlations of metrics with \textbf{human judgement}.
 We report the maximum over the error types in Table~\ref{tab:taxonomy-simple}.
 All metrics are measured \textbf{reference-free} on $(\bm{s},\bm{h})$. The highest two correlations in each column are \textbf{bolded} and \underline{underlined}, respectively. Correlations that are not significant (\textit{p-value} $\ge 0.05$) are omitted when aggregating, and "-" denotes an absence of any significant correlation.
 Breakdown of all baseline and \ourmodel~ metrics is shown in App.~\ref{app:human-experiments}.
 }
	\label{tab:corr-human-unsup-simple-max}
	\begin{center}
	   \scalebox{0.75}
    {
 \begin{tabular}{@{}lccccc@{}}
    \toprule
          & \hyperref[tab:corr-human-unsup-DROP]{\textbf{DROP}}
          & \hyperref[tab:corr-human-unsup-GSM8K]{\textbf{GSM8K}}
          & \hyperref[tab:corr-human-unsup-ESNLI]{\textbf{ESNLI}}
          & \hyperref[tab:corr-human-unsup-COSMOS]{\textbf{COSMOS}}
          & \hyperref[tab:corr-human-unsup-SEMEVAL]{\textbf{SemEVAL}} \\
          \midrule 
Rouge-L & 0.278 & 0.252 & 0.557 & -0.441 & -0.478  \\
 BLEURT & 0.328 & 0.256 & 0.541 & 0.218 & -0.356  \\
 BERTScore & 0.275 & 0.235 & 0.590 & -0.420 & -0.295  \\
 \bartscore & -0.835 & -0.546 & 0.549 & -0.544 & -  \\
 \bartscoreParaCnn & -0.665 & - & 0.482 & -0.186 & -  \\
 \bartscoreFineTuned & -0.642 & - & 0.255 & -0.207 & -  \\
 PRISM & -0.733 & -0.455 & 0.580 & -0.376 & -  \\
 CTC-Relevance & 0.333 & -0.371 & 0.334 & - & -0.349  \\
 CTC-Consistency & 0.462 & -0.174 & \underline{0.647} & 0.275 & -0.301  \\
    \midrule
  \ourmodelsa & 0.578 & 0.392 & 0.521 & 0.555 & 0.337  \\
  \ourmodelsa$^1$ & 0.790 & 0.500 & \textbf{0.799} & 0.638 & 0.485  \\
   \ourmodelsa$^2$ & 0.578 & 0.392 & 0.599 & 0.555 & 0.337  \\
      \midrule
   \ourmodelss & \textbf{0.824} & \underline{0.514} & 0.530 & 0.593 & 0.411  \\
   \ourmodelss$^1$ & 0.791 & 0.471 & 0.507 & \underline{0.642} & 0.508  \\
  \ourmodelss$^2$ & \underline{0.799} & \textbf{0.638} & 0.531 & \textbf{0.658} & \textbf{0.535}  \\
     \midrule
  \ourmodelli & 0.584 & 0.345 & 0.531 & 0.444 & 0.372  \\
     \midrule
  \ourmodellc & 0.205 & -0.184 & 0.447 & -0.212 & \underline{0.517}  \\
    \bottomrule
\end{tabular}
    }
	\end{center}
\end{wraptable}

\textbf{Meta-Evaluations on Human Judgement Datasets}.
\label{human-experiments}
Table~\ref{tab:corr-human-unsup-simple-max} reports a summary of meta-evaluation of \ourmodel~ metrics comparing against baselines on human judged datasets. The correlations are measured based on the presence of a particular error from Table~\ref{tab:taxonomy-simple} and we report the highest correlation across all error types within each dataset.
We observe that: (1) on all tasks, \ourmodel~metrics outperform all other baselines when evaluated as reference-free;
(2) overall, \ourmodel~yields considerably better correlations, which indicates that \textit{step-by-step} reasoning generations can be more effectively evaluated with \ourmodel.
In general, most correlations with human judgements are moderate when compared to the synthetic correlation scores, indicating that step-by-step reasoning evaluation is among the cognitively hard tasks for neural models \citep{deutsch2022re}.
Interpretable metrics such as \ourmodel~can provide better information about a model's reasoning skills, thus future work should improve such metrics on aligning with human judgments. In App.~\ref{app:human-experiments}, we show fine-grained experimental analysis per each human labeled dataset. Specific examples showcasing ~\ourmodel~scoring abilities are summarized in Table~\ref{tab:examples}.
 
\section{Analysis}
\label{analysis}
\begin{wrapfigure}{r}{0.3\textwidth}
    \centering
    \scalebox{0.5}
    {
    \includegraphics{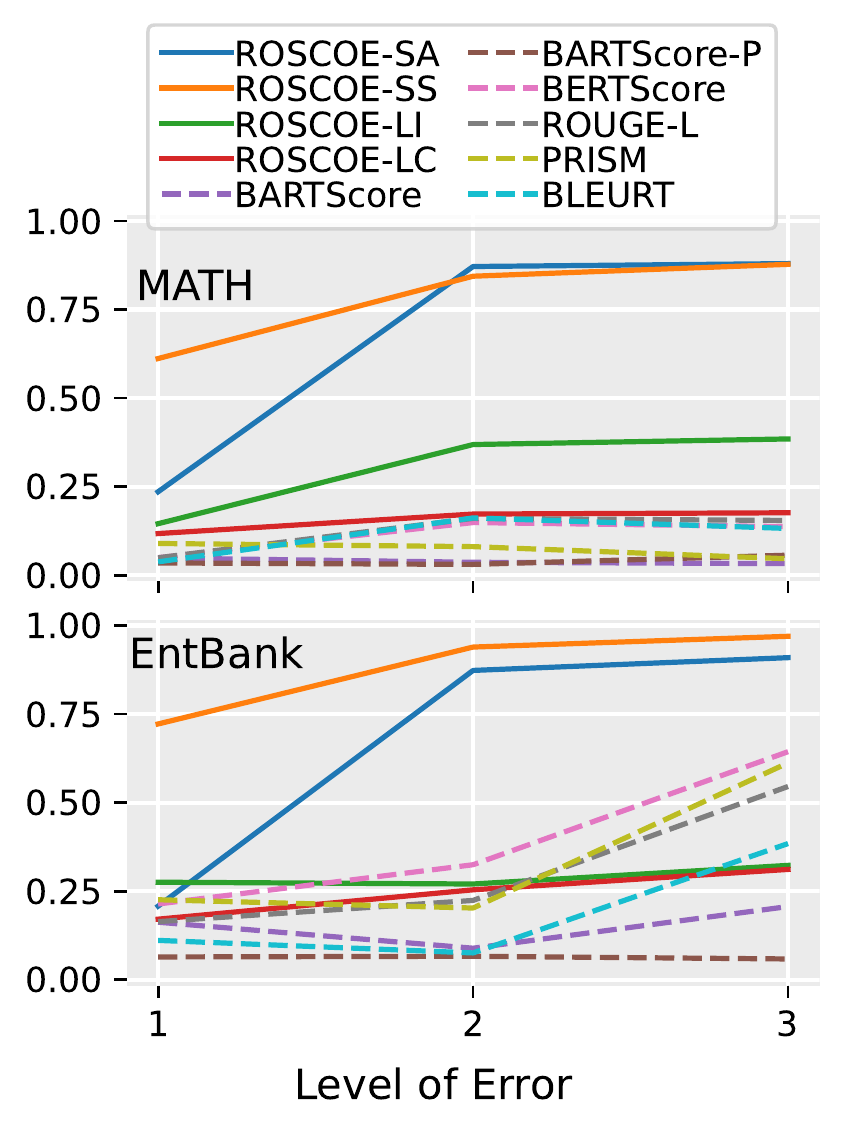}
        }
    \caption{\footnotesize Sensitivity of selected metrics on Somers' \textit{D} by injecting levels of error into reasoning steps.}
    \label{fig:sensitivity-simple}
    \vspace{-0.1em}
\end{wrapfigure}

\paragraph{How sensitive are ~\ourmodel~metrics against level of errors?}
\label{sensitvity}
To evaluate how well metric values match human assessment of reasoning, we measure sensitivity to the level of errors. We perturb sentences in the MATH (arithmetic) and EntailmentBank (deductive reasoning) diagnostic datasets (similar to $\S$~\ref{experimentsetup}) and inject different levels of errors into the reasoning text. Using randomly selected perturbation types, we construct up to a maximum of 3 perturbations per instance. We measure
the correlation (Somers' $D$) between the reasoning
inconsistency level 1, 2, 3 of the reasoning steps (i.e., the
number of injected errors) and the metric score. Fig.~\ref{fig:sensitivity-simple} illustrates the results averaged over different perturbations. 

We expect the metrics correlate with humans better when the level of errors is high. Both semantic alignment of the reasoning \ourmodelsa~, and the semantic similarity metrics \ourmodelss~ show consistent behavior on both datasets, while baseline metrics fluctuate with low correlations.
Baseline metrics perform better on EntailmentBank. 
On MATH, \ourmodellc~ and the baseline metrics 
show minimal impact, which can be that some of the perturbations applied on the MATH dataset (e.g., RandomOperation, or ShuffleNumbers) are harder to detect with language model based (BARTScore) and NLI model based (\ourmodellc) metrics. 

\begin{figure}[!h]       
   \vspace{5px}
    \mbox{\includegraphics[height=.18\textheight]{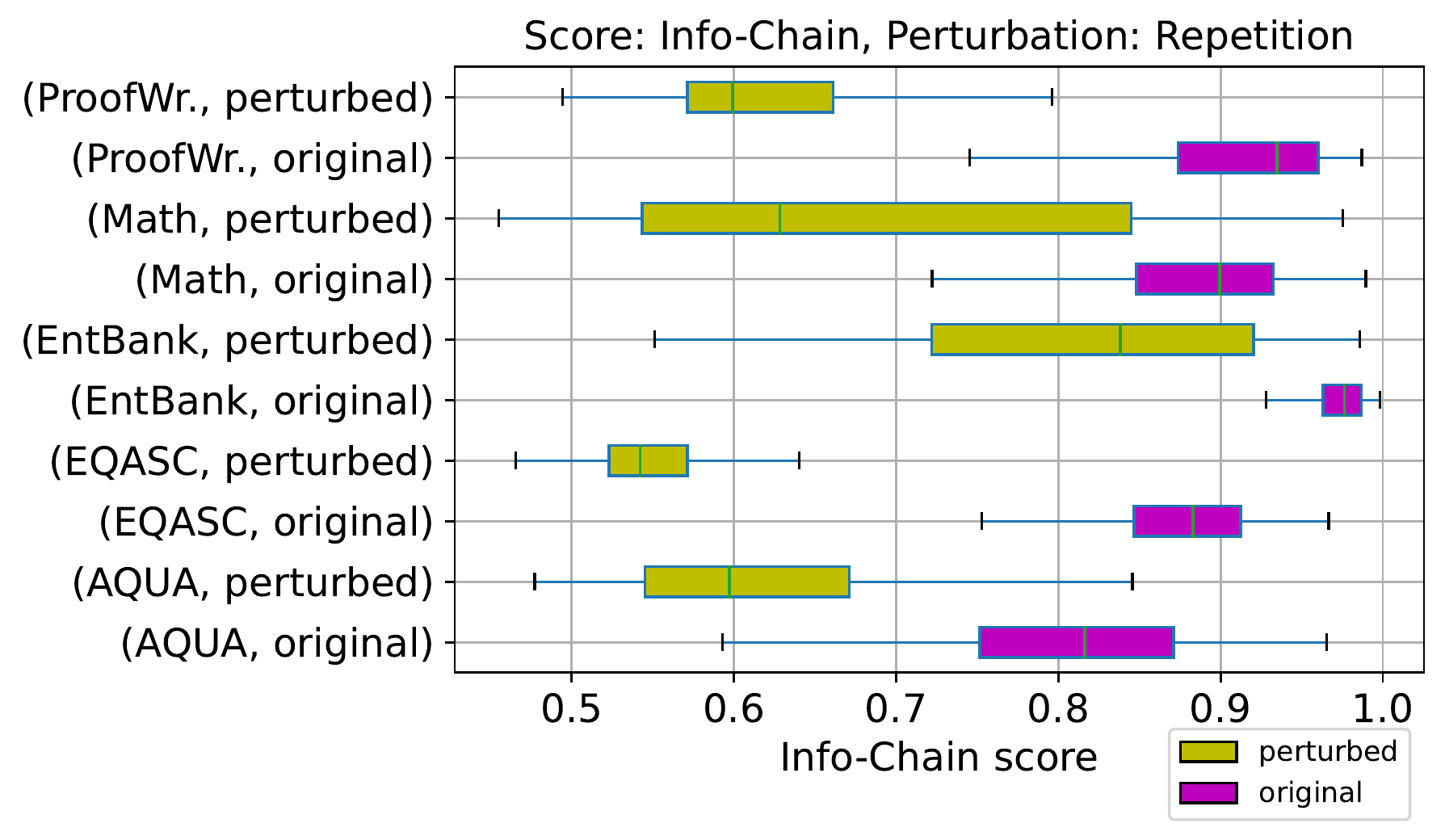}}   
    \hspace{-8px}
    \mbox{\includegraphics[height=.18\textheight]{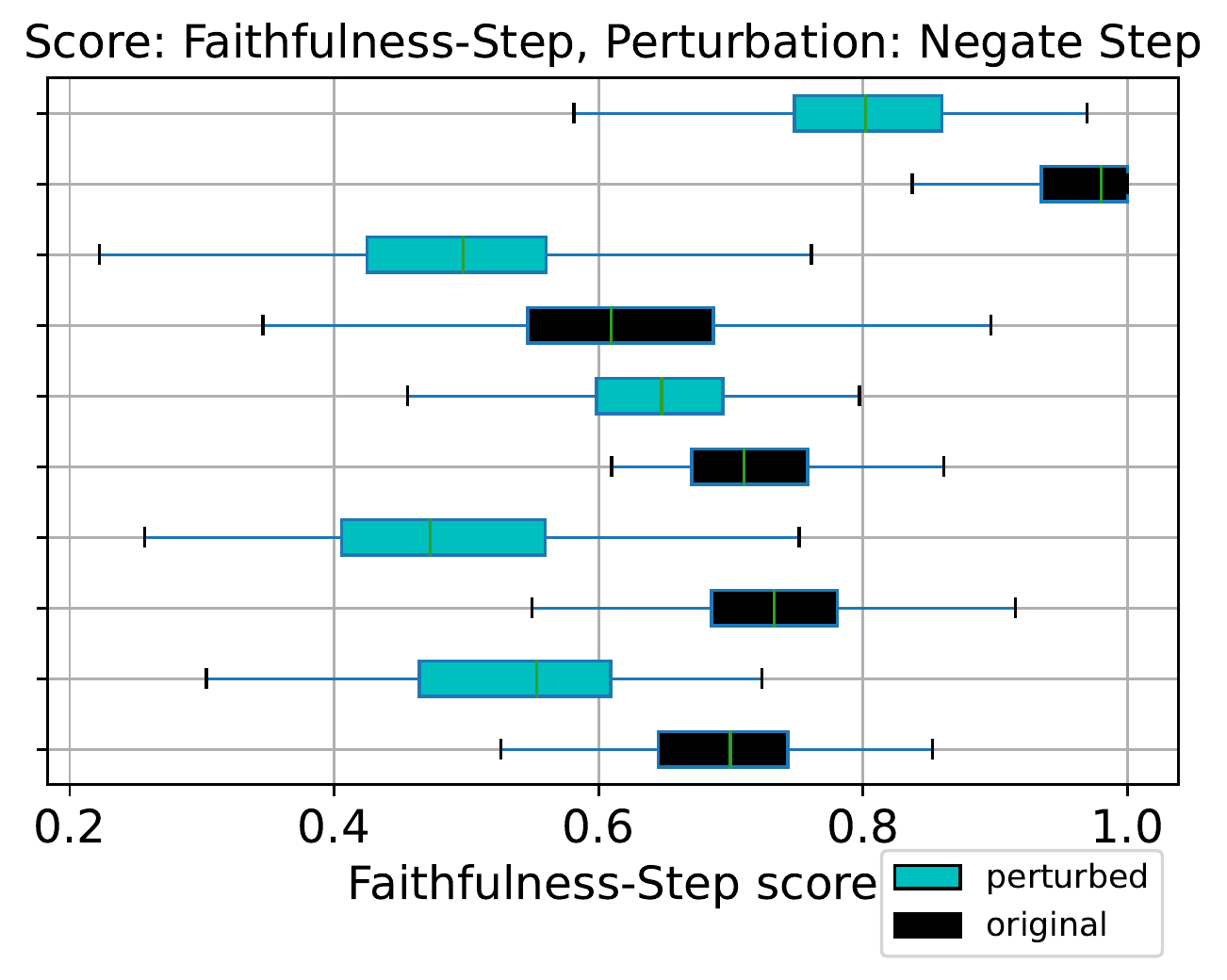}}
    \hspace{-8px}
    \mbox{\includegraphics[height=.18\textheight]{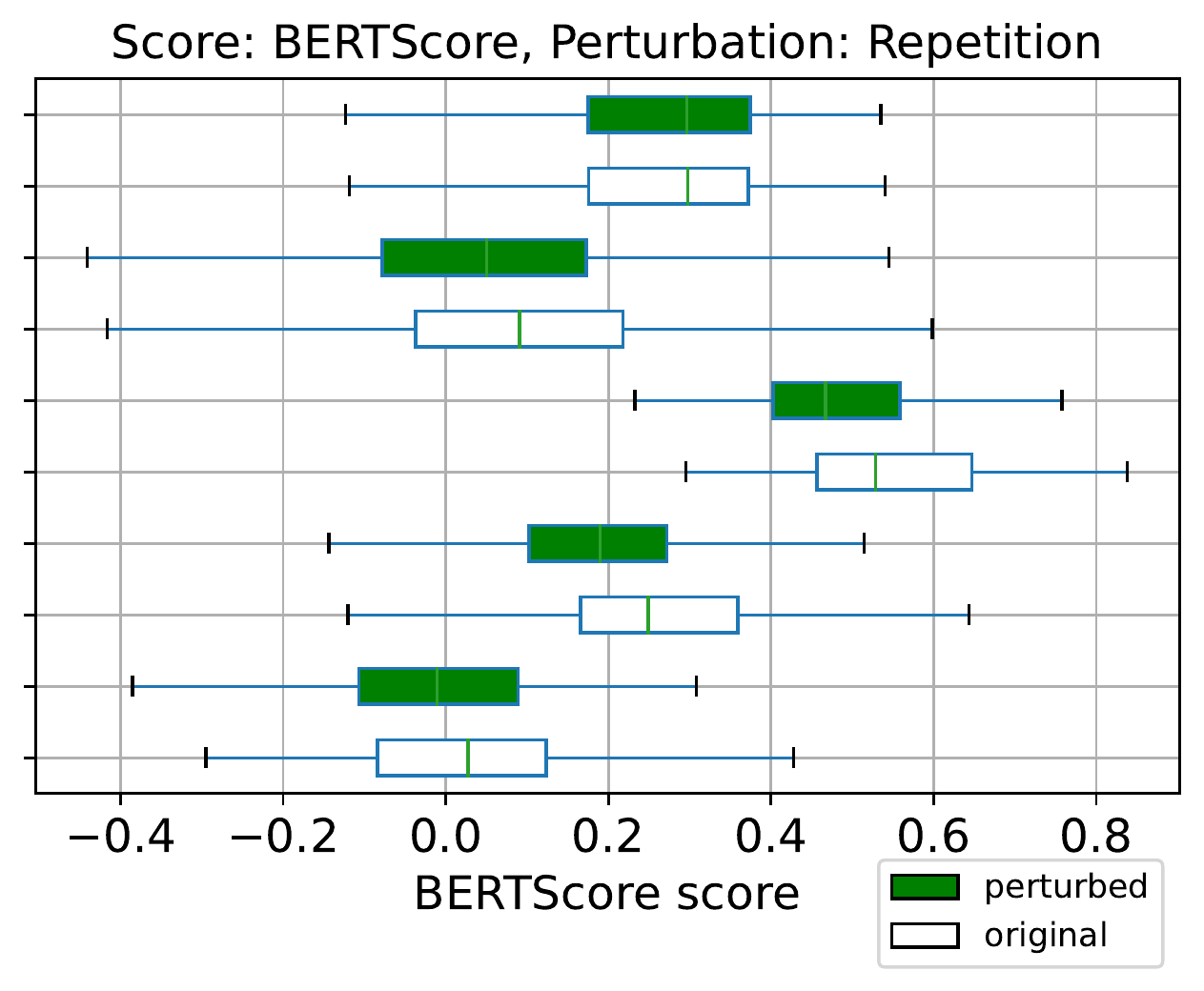}}
    \caption{\footnotesize Box-and-whisker plots of interquartile ranges of scores, for perturbations and reference-free metrics with strong Somers' $D$ values. Scores are split by dataset and perturbation use. While interquartile ranges separate well by perturbation use within a single dataset, there is overlap across datasets. This shows the drift of neural scores across datasets and applies to both \ourmodel~(left, center) and strong baselines (right).} 
    \label{fig:score_range}
\end{figure}

{
\parfillskip=0pt
\parskip=0pt
\par}
\textbf{What does \ourmodel~illuminate about scores across errors and tasks?}
\label{score_range_analysis}
For an ideal scorer based on ease of use, it would be possible to pick a set of fixed thresholds that had error discrimination power across datasets. However, we show that this dataset-agnostic ideal is currently not possible and an issue endemic across scores, including baselines.   
We study which metrics correlate strongly with which perturbations, with a focus of consistency across datasets. From this, we plot the interquartile ranges for strongly correlated metric and perturbation pairs. We show a sample of these in Fig.~\ref{fig:score_range}, though find that the trends generally hold across metrics and perturbations (see Fig~\ref{fig:score-perturb}). We note that within a given dataset, scores are well separated: the \textit{perturbed} version of a dataset for a given score and perturbation type shows little interquartile overlap with the \textit{original} version. However, this does not hold across datasets -- e.g., in (Score: Info-Chain, Perturbation: Repetition), if one were to set a detective threshold for the Repetition perturbation based off EntBank (around $0.95$), it would mark almost all values of EQASC as perturbed, even non-perturbed samples. This shows the challenge of using metrics for classification
without calibration for drifts in both mean and variance across datasets, even if a metric generally correlates well with detecting a given error. 

\section{Conclusion}
In this paper, we introduce \ourmodel, a new suite of interpretable, unsupervised metrics that enables evaluation of step-by-step reasoning generations of LMs when no golden reference generation exists. We present a taxonomy of reasoning errors used to generate and evaluate our metrics. Experimental results, from evaluating on both synthetic and human-labeled datasets exhibiting multiple types of reasoning (commonsense, arithmetic, and logical inference, etc.), demonstrate superior performance compared to prior semantic and lexical similarly based baseline metrics for text generation. Our analysis shows improved capability in evaluation of reasoning exhibiting nuances, such as factual and logical errors in step-wise decisions.  

\section*{Ethics Statement}
\label{app:ethics}
Explainability builds transparency and trust for users, eases bug-fixing and shortens
improvement cycles for metric designers, and will be required by law/regulations for AI systems
to be applied to large-scale, high-stakes domains. In this context, we hope our work will catalyze
efforts on the topic of explainable evaluation metrics for language model rationale generations. 
We should mention that our evaluation metrics do not monitor the explanations from integrity or bias perspectives. Our work also uses five human expert annotators and in the annotation process, annotators need to rate the model generated candidate rationals. While the model-generated explanations can produce potentially unsafe content, the datasets for annotations include domains related to logical and arithmetic concepts and general commonsense knowledge. The anecdotal consensus was that the generations were safe and didn't include biased statements.

\section*{Reproducibility Statement}
To ensure the reproducibility of our empirical results, we will open source our code to Github, which will contain: instructions for installing the virtual environment, data preprocessing, all score generation and correlation scripts (both for \ourmodel~and baselines), and trained embedding models. Detailed explanation of all the finetuned models and metrics
are given in the main paper as well as in the Appendices. We will also release all the diagnostic and human judgment datasets used in our experiments. 

\bibliography{iclr2023_conference}
\bibliographystyle{iclr2023_conference}

\clearpage
\section*{Appendix}
\appendix


\section{Limitations}
\label{app:limitations}
Our study is the first initial step that investigates the evaluation of the step-by-step reasoning produced by large language models. Our taxonomy (in Table~\ref{tab:taxonomy-simple}) covers several reasoning errors and we designed our metrics to evaluate a spectrum of criteria including the ones in the taxonomy. Even though we cannot say we cover all possible reasoning errors, our metrics are generic enough, work on natural language rationales, and consider the alignment with the input context and the generated explanation. Nevertheless, we believe our study can spur others to investigate different reasoning errors and use our code and datasets as templates to extend further.

Due to the extensive analysis needed to thoroughly test and communicate the ability of our proposed metrics to capture reasoning errors, we decided to leave some follow-up questions, such as the application of these metrics for improving downstream task performance, for future exploration.

\section{Few-shot Prompting Examples (Cont. from \S~\ref{intro})}
\label{app:intro}
Below is the 2-shot example we used to generate the explanations from GPT-3 as we show in the Fig.~\ref{introfigure}.

\begin{center}
\scalebox{0.90}
	    {
\fbox{\begin{minipage}{40em}
I will explain how to make logical inferences: \\

\textbf{Context:} John wakes to the sound of his children giggling outside the door. They bust in and say "Happy Father's Day!" with a plate full of food. 2022 has been a great
year. 

\textbf{Question:} Can we infer that John was born before the inauguration of Barack Obama?

\textbf{Explanation}: We can infer that John is a father because of "his children". We can infer that the current year is 2022. Barack Obama was inaugurated in 2009, so it has been
about 13 years since Obama was inaugurated. John's children are old enough to speak, so if John were born after Obama's inauguration, he would have been younger than
10 when he had children. So, we can infer that John was born before Obama's inauguration. The answer is "yes".

\paragraph{Context:}Sam sees a piano in the subway stations. Laughing, his friends prod him to play, saying "Remember how amazing your rendition of Chopsticks was at the Christmas
party?" 

\textbf{Question}: Can we infer that John majored in piano at Juilliard?

\textbf{Explanation}: Juilliard is a top school for music, so someone who majored in piano at Juilliard would be an expert pianist. "Chopsticks" is a very simple song to play on
the piano. His friends were therefore likely being sarcastic when they told him to play, which would imply that Sam was not an expert piano player. So it is unlikely
that Sam majored in piano at Juilliard. The answer is "no".
\end{minipage}}
}
\end{center}

Table~\ref{tab:promptexample1} illustrates sample if in-context examples we used as prompts to obtain step-wise explanations. 
\begin{table}[!b]
\vspace{-10pt}
\caption{\footnotesize Example prompts used to generate \textit{step-by-step} explanations using GPT-3.  
}
\begin{center}
\renewcommand{\arraystretch}{1.2}
\scalebox{0.80}
	    {
\begin{tabular}{@{}p{460pt}}
\toprule
(Used for ESNLI dataset) \par
Explain if premise entails hypothesis. \par
\textbf{Premise}: Tomato sauce (also known as Neapolitan sauce, and referred to in Italy as Salsa di pomodoro) refers to any of a very large number of sauces made primarily from tomatoes, usually to be served as part of a dish (rather than as a condiment). Tomato sauces are common for meat and vegetables, but they are perhaps best known as sauces for pasta dishes. \par
\textbf{Hypothesis}: Tomato sauce is only rarely used to top pasta. \par
\textbf{Explanation}: Salsa di pomodoro is made of peppers. The answer is No. 
\par\par 
\ \ \par
\textbf{Premise}: Kirk Johnson (born 1960) is an American scientist, author, curator, and museum administrator, and is currently serving as Director of Smithsonian's National Museum of Natural History. Dr. Johnson is the host of the PBS Nova series, \"Making North America\", which is a three-part series that describes the shaping of North America, which aired on November 4, 11 and 18, 2015. \par\par
\textbf{Hypothesis}: johnson hosted Making North America when he turned 50. \par
\textbf{Explanation}: Johnson was born in 1960 and hosted Making North America in 2015, so he was 55, not 50. The answer is No. \par
\ \ \par
\textbf{Premise}: A person rides his bicycle in the sand beside the ocean.\par
\textbf{Hypothesis}: A person is on a beach.\par
\textbf{Explanation}: A person rides beside the ocean means is on a beach if there is sand beside the ocean. The answer is Yes.\\
\midrule
(used for SEMEVAL dataset) \par
Explain if premise entails hypothesis. \par
\textbf{Premise}: I went into my bedroom and flipped the light switch. Oh, I see that the ceiling lamp is not turning on. It must be that the light bulb needs replacement. I go through my closet and find a new light bulb that will fit this lamp and I place it in my pocket. I also get my stepladder and I place it under the lamp. I make sure the light switch is in the off position. I climb up the ladder and unscrew the old light bulb. I place the old bulb in my pocket and take out the new one. I then screw in the new bulb. I climb down the stepladder and I place it back into the closet. I then throw out the old bulb into the recycling bin. I go back to my bedroom and turn on the light switch. I am happy to see that there is again light in my room. Which room did the light go out in?\par
\textbf{Hypothesis}: Bedroom
\par
\textbf{Label}: Correct \par
\textbf{Explanation}: The answer is bedroom, because the light went out in the bedroom. The person replaced the light bulb in the bedroom. \par
\ \ \par
\textbf{Premise}: The weather was so nice today that I decided to have a barbecue. I called up some of my friends to invite them over. I set up some chairs and tables outside for my friends to sit around. I then took out some hot dogs from my fridge and brought them over to my barbecue. My barbecue is a gas barbecue, so all I had to do was press the knob and turn it to ignite it. Once I got a little fire going I put my hot dogs on the grill over it to cook. I closed the lid and left my hot dogs to cook for about ten minutes. After ten minutes I decided to check on them to see if they were done. They looked like they were cooked, so I put them all on a plate to serve to my friends. I went to my kitchen and got hot dog buns, ketchup, mustard, and relish to serve on top of the hot dogs. I laid out all of the food and condiments on the table so my friends could help themselves. We ate the hot dogs and relaxed around the table. Why are we having the barbecue, to celebrate an occasion, or just because we want to?
\par
\textbf{Hypothesis}: Just because they want to. 
\par
\textbf{Label}: Correct
\par 
\textbf{Explanation}: The answer is correct, because the weather was nice. They decided to have a barbecue. There is no mention of any special occasion. \par
\ \ \par
\textbf{Premise}: Another Nurse can in to look at me , took one look at my rash and said OMG and ran out of the room . At that point I started crying , regretted coming to the hospital alone and started wheezing again . Then the doctor came in , her eyes were all bugged out . What will I do after the doctor examines me ?\par 
\textbf{Hypothesis}: I will talk with my friends .
\par 
\textbf{Label}: Incorrect
\par
\textbf{Explanation}: The premise is about the nurse's reaction to the patient's rash, and how the patient feels. The hypothesis does not follow up from this premise correctly. Instead, it talks about talking with friends. \\
\bottomrule
\end{tabular}
}
\end{center}
  \label{tab:promptexample1}%
\end{table}

\newpage
\section{Taxonomy of Reasoning Errors (Cont. from \S~\ref{taxonomy})}
\label{app:taxonomy}
To gain deeper insights into the types of
reasoning errors introduced by LLMs while explaining their decisions, we propose a new taxonomy of generic reasoning errors for language problem solving. Specifically, we sampled from the training portions of the logical inference and commonsense reasoning datasets, and prompted GPT-3 with reasoning explanations using prompts similar to App.~\ref{app:intro}. 
We used task specific in-domain examples for prompting. We also analyzed model generated explanations shared in \citeauthor{wei2022chain} (\citeyear{wei2022chain}). 
We then manually looked into each explanation and identified potential errors that are inconsistent with the source, question or the prompt and within the reasoning chain. Some tasks
require a model to classify the logical relationship
between premise and a hypothesis, others are question and answering tasks. We adjusted our context and prompts according to the type of the task.

Our reasoning error taxonomy is summarized in Table~\ref{tab:apptaxonomy}. It contains types of errors concerning an overall chain or an individual step. Specifically, the \textit{chain-level} \textit{coarse-grained} evaluations of the overall reasoning chain deals with overall quality of the step-by-step thinking, coherence, consistency of the explanation within itself, and consistency with the context, etc. On the other hand the \textit{step-level} \textit{fine-grained} evaluations focus on the consistency of a reasoning step with the previous steps, if a step conveys new and supporting information over the previous steps, factuality or logical inference issues. We use these error categories to construct diagnostics datasets with perturbed errors as well as human judged datasets of reasoning errors. In the taxonomy, we indicate *-step level errors to differentiate from the chain level error types. 

\begin{table}[p]
\vspace{-10pt}
\caption{\footnotesize Taxonomy of Step-by-Step Reasoning Errors. Errors used for perturbations in constructing the diagnostic datasets (Diag.) and for human annotation (Human) of the model generated reasoning chains are also marked. (Cont. from Table~\ref{tab:taxonomy-simple}.
}
\begin{center}
\renewcommand{\arraystretch}{1.2}
\scalebox{0.80}
	    {

\begin{tabular}{@{}p{60pt}p{140pt}p{240pt}c@{}c@{}}
\toprule
\textbf{Error Type}
& \textbf{Description} \ \
& \textbf{Example} \ \
& \textbf{Diag.} \ \ 
& \textbf{Human}
\\ \midrule

Grammar 
& Faulty, unconventional, or controversial grammar usage
& \textbf{Reference}: He chews 4 pieces of gum a day so over 30 days he will chew 4*30 = <<4*30=120>>120 pieces of gum. \par
\textbf{Model Expl}: \textcolor{red}{He eats 4 pieces of gum a day and lasts 30 days} so he eats 4*30 = <<4*30=120>>120 pieces of gum. 
& \checkmark & \checkmark \\\cline{1-5}

Factuality
& Information about object (i.e. quantity, characteristics) or a personal named entity does not match information provided in the question
& \textbf{Context}: Anakin caught 10 fish. \par
\textbf{Model Expl}: \textcolor{red}{Locsin} caught 10 fish. 
& \checkmark & \checkmark \\\cline{1-5}

Hallucination
&  Information is not provided in the problem statement and is irrelevant or wrong
& \textbf{Context}: The basketball team went to the steakhouse to eat dinner.  The first player ate a 6-ounce steak.  The second player ate beef tips, containing 8 beef tips, each an ounce in size.  The third player ate a one-pound steak.  And the fourth and fifth players ordered vegetarian meals.  In total, how many ounces of meat were consumed by the team? \par
\textbf{Model Expl}: The fourth and fifth players ordered \textcolor{red}{vegetarian meals, for a total of 2 ounces of meat}. 
& \checkmark & \checkmark \\\cline{1-5}

Redundancy
&  Explanation contains redundant information, which even though might be factual, is not required to answer the question
& \textbf{Context}: Bob is cold. If something is quiet then it is blue. All blue things are round. If something is round and smart then it is cold. Question: Bob is cold. \par
\textbf{Model Expl}: \textcolor{red}{If something is round and not smart then it is blue. All blue things are round}. Answer: true.
&  & \checkmark \\\cline{1-5}

Repetition
& Step paraphrases information already mentioned in previous reasoning steps
& \textbf{Context}: Parker chews 4 pieces of gum a day.  A pack of gum has 15 pieces of chewing gum per pack.  How many packs of gum will he need to last him 30 days?
\par \textbf{Model Expl.}: He chews 4 pieces of gum a day and there are 30 days in a month so he chews 4*30 = 120 pieces of gum a month. \textcolor{red}{Parker chews 120 pieces of gum a month.} He needs 120 pieces of gum a month and a pack of gum has 15 pieces of gum so he will need 120/15 = 8 packs of gum.
& \checkmark & \checkmark \\\cline{1-5}

Missing step
&  The content of the generated reasoning is incomplete and lacks required information to produce the correct answer. If this missing steps are added, the model would produce the correct answer
& \textbf{Context}: A farm has 10 2-legged animals and 15 4-legged animals. If all animals are healthy, how many pairs of animal legs in total are on the farm? \textbf{Answer}: 40
\par \textbf{Model Expl.}: 10 2-legged animals have 10*2=20 legs. 15 4-legged animals have 15*4=60 legs. In total on the farm, there are 20+60=80 legs.\textbf{Answer}: 80 \textcolor{red}{missing step: Total number of pairs is 80/2=40}
& \checkmark & \checkmark \\\cline{1-5}

Coherency
& Steps contradict each other or do not follow a cohesive story
& \textbf{Context}: If the reactor is intact, then it is safe. 
If the reactor is safe, then no graphite is outside it. \par
\textbf{Model Expl.}: The reactor is intact. \textcolor{red}{All graphite is inside of it. There are some graphite outside of the the reactor.}
& \checkmark & \checkmark \\\cline{1-5}

Commonsense
&  Model lacks relations that should be known from general knowledge about the world (i.e. how to compute velocity, how many inches in one foot, all ducks are birds, etc.)
& \textbf{Question}: Jordan has 2 children who wear diapers.  Each child requires 5 diaper changes per day. Jordan's wife changes half of the diapers. How many diapers does Jordan change per day? \par
\textbf{Model Expl.}: Jordan has 2*5=10 diapers to change per day. Jordan's wife changes 10/2=5.0 diapers per day. For both children, Jordan changes \textcolor{red}{10+5=15} diapers per day. 
&  & \checkmark \\\cline{1-5}

Arithmetic
&  Error in math calculations
& \textbf{Reference}: If a truck was carrying 20 tons of fertiliser packed in bags, the total number of bags in a truck is 20*20 = 400 bags \par
\textbf{Model Expl}: If a truck was carrying 20 tons of fertiliser packed in bags, the total number of bags in a truck is \textcolor{red}{20*20 = 40} bags
& \checkmark & \checkmark \\\cline{1-5}

\bottomrule
\end{tabular}
}
\end{center}
  \label{tab:apptaxonomy}%
\end{table}
\newpage
\section{\ourmodel~Metrics Details (Cont. from $\S$\ref{roscoethemetric})}
\label{app:roscoe}
\begin{figure}[h]
\caption{\textbf{Reasoning alignment} illustrating the measurement of the Faithfulness-Step and Faithfulness-Token semantic alignment scores. $\vh = \{h_1, h_2\}$ is a hypothesis chain with tokens $\{h_{1,1}, h_{1,2}, h_{1,3}, h_{2,1}, h_{2,2}\}$, and $\vs=\{s_1, s_2, s_3\}$ is a context with tokens $\{s_{1,1}, s_{2,1}, s_{2,2}, s_{2,3}, s_{3,1}, s_{3,2}, s_{3,3}\}$. Alignment scores from hypothesis to context are highlighted, and alignment scores from context to hypothesis are underscored. The reasoning alignment combines token and step level similarities where each alignment value (cell) is the cosine similarity
and explicitly measures the grounding of the token and step-wise reasoning with respect to the source text.}
\label{fig:alignment}
\centering
\includegraphics[width=15cm]{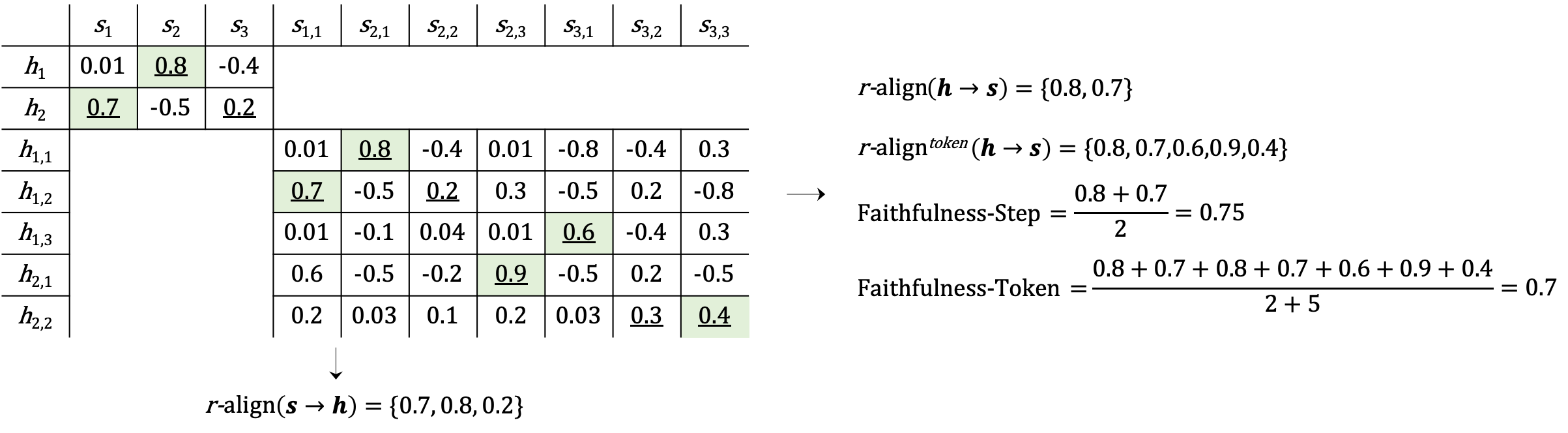}
\end{figure}

\ourmodel~metrics are constructed under four categories: semantic alignment, semantic similarity, logical inference, and logical coherence. The details of each metric is explained in $\S$\ref{roscoethemetric}. At the core of \ourmodel~ semantic alignment metrics is the reasoning alignment score, which we designed to measure the grounding of step-by-step reasoning with respect to the source text. Fig.~\ref{fig:alignment} illustrates the reasoning alignment. 

\vspace{0.3cm}

The variation of scorers of the \ourmodel~shares some similarities, thus we explain them here: 

\textbf{BARTScore}~\citep{bartscore} claims that more high level text can be generated using sequence to sequence model. It can support different evaluation perspectives such as factuality (by evaluating from source to hypothesis) or informativeness (by evaluating from both directions between reference and hypothesis). \bartscore~ is used to measure the probability of generated text from a source text $x$ to a target set $y$:
\begin{equation}
    \bartscore~ = \textstyle \sum^m_{t=1}w_t\log p(y_t|y_{<t},x,\theta)
\end{equation}
\bartscore introduce two variations: (1) finetuning, in which the BART model is finetuned on the task specific dataset to make the pre-training domain closer to the evaluation domain. (2) prompting, in which a task specific textual prompt is appended to the source $x$ to get the $y$. In our experiments we compare the the \bartscore baseline and one with the prompting variant \bartscoreParaCnn to compare in the experiments.  

\textbf{CTC} (Compression, Transduction, and Creation)~\citep{ctc}, is a suite of metrics that unifies different perspectives of different tasks (e.g, summarization, style transfer, or text rewriting) into information alignment, which measures weather the information in one generation component is grounded in another. The information alignment is defined as follows: let $x$ (e.g, dialog context) be the source input, $c$ (e.g., external world knowledge) be some additional context, and $y$ be the generated output text (e.g., generated response). The alignment is measured on token level and it is measured as the vector of scores:
\begin{equation}
    align(a\rightarrow b) = \left<\alpha_1, \cdots,\alpha_N \right>
\end{equation}
where each score $\alpha_i$ indicates confidence that the \textit{n}-th token in $a$ aligns with the whole sentence $b$. Using the information alignment they define a list of metrics to evaluate text for different tasks. In our experiments we use two of these metrics that are closer to \ourmodel: the \textit{Relevance} (CTC Relevance), which measures the consistency of the generated text with the source and its balanced between the reference, and the \textit{Consistency} (CTC Consistency) which deals with the faithfullness of the generated text to the input context by the alignment between the two.

\newpage
\section{Experimental Setup Details (Cont. from $\S$~\ref{experimentsetup})}
\label{app:setup}
\subsection{Diagnostic Datasets} 
\label{app:diagnosticdatasets}

In the following we present details of each diagnostics dataset used in our work. Table~\ref{tab:dataset-appendix} illustrates how each dataset is used in our experiments. StrategyQA dataset is only used to finetune the SimCSE embeddings model, because it contains reference reasoning chains in train and validation partitions, but not in the test partition.  The rest of the six diagnostic datasets are used for sentence embedding model finetuning, and evaluating our models as presented in the experiments results. All datasets with examples are summarised in  Table~\ref{alldatasetssamples-diaognostic}. 
\vspace{0.5cm}
\begin{table*}[h!]
  \centering
  \footnotesize
    \renewcommand{\arraystretch}{1.2}
      \caption{\label{tab:dataset-appendix} Summary of datasets used in our work. Reasoning Chain represent whether it contains human written golden step-wise reasoning explanation. Type indicates whether it is used for constructing Diagnostic or Human judged datasets. Train/Val./Test indicate whether the dataset is used for training, validation and/or testing. StrategyQA dataset is only used for finetuning SimCSE embedding model. 
      }
\begin{tabular}{lcccccr}
\toprule
Dataset & Reasoning & Type & Train & Val. & Test & Annotated  \\
& Chain &  & &  & & Instances  \\
\midrule
EntailmentBank~\citep{entalmentbank2021} & \cmark & Diagnostic, Finetuning &  \cmark& \cmark & \cmark& 1,840 \\
ProofWriter~\citep{tafjord-etal-2021-proofwriter}& \cmark & Diagnostic, Finetuning &\cmark & \cmark &\cmark & 272,430 \\
MATH~\citep{hendrycksmath2021} & \cmark & Diagnostic, Finetuning & \cmark& \cmark & \cmark& 12,500 \\
ASDIV~\citep{miao-etal-2020-diverse} & \cmark & Diagnostic, Finetuning & \cmark& \cmark & \cmark& 2,305 \\
AQUA~\citep{liang-etal-2018-meaning} & \cmark & Diagnostic, Finetuning & \cmark& \cmark & \cmark& 97,975 \\
EQASC~\citep{aggarwaletal2021ecqa} & \cmark & Diagnostic, Finetuning & \cmark& \cmark & \cmark& 9,060\\
StrategyQA~\citep{geva2021did} & \cmark & Finetuning & \cmark& \cmark &\xmark  & 2,290 \\
DROP~\citep{dua-etal-2019-drop} & \xmark & Human judged &\xmark  & \xmark  & \cmark & 210 \\
GSM8K~\citep{cobbe2021gsm8k}  & \cmark & Human judged & \xmark & \xmark  & \cmark & 200 \\
ESNLI~\citep{esnli}& \cmark & Human judged &\xmark  & \xmark  & \cmark& 151 \\
CosmosQA~\citep{huang-etal-2019-cosmos} & \xmark & Human judged &\xmark  & \xmark  & \cmark & 195  \\
SemEval~\citep{Ostermann2018SemEval2018T1} & \xmark & Human judged &\xmark  & \xmark  & \cmark & 209  \\
\bottomrule
\end{tabular}
\end{table*}
\begin{table}[h!]
  \centering
\caption{\footnotesize We show instances from seven of the \textbf{Diagnostics Datasets} here. (Continue from $\S$\ref{datasets}).
}
  \vspace{-10pt}
\label{alldatasetssamples-diaognostic}
    \scalebox{0.75}{
  \begin{tabular}{@{}lp{230pt}p{250pt}@{}}
    \toprule
    \textbf{Dataset} & \textbf{Instance} & \textbf{Reference Step-by-Step Solution}  \\
    \midrule
\textbf{EntBank} & Earth is a kind of celestial object. Stars appear to move relative to the horizon during the night. A star is a kind of celestial object celestial body. The earth rotating on its axis causes stars to appear to move across the sky at night. Apparent motion is when an object appears move relative to another object 's position. \par \textbf{Question}: How does the appearance of a constellation change during the night? \par \textbf{Hypothesis:} Solve the following entailment problem: "Earth is a kind of celestial object. During the night stars appear to move" & 
\textbf{Step1}: earth is a kind of celestial object Its position appears \par
\textbf{Step2}: a star is a kind of celestial object to shift relative / celestial body to the horizon.\par
\textbf{Step3}: apparent motion is when an object appears to move relative to another object 's position \par
\textbf{Step4} Therefore apparent motion of stars is when stars appear to move relative to earth's position 
\textbf{Step5}: The earth rotating on its axis causes stars to appear to move across the sky at night \par
\textbf{Step6}: Therefore the earth rotating on its axis causes apparent motion of stars \par
 \textbf{Step7}: Stars appear to move relative to the horizon during the night \par
 \textbf{Step8}: Therefore the earth rotating on its axis causes stars to move relative to the 
  horizon during the night. \\
  \midrule
     \textbf{Proof}\par \textbf{Writer} & \textbf{Facts}: 
     The cow is not big. The cow is not green. The lion eats the tiger. The lion sees the cow. The lion visits the cow. The lion does not visit the squirrel. the lion visits the tiger. The squirrel is big. The squirrel is round. The tiger is not green. The tiger does not see the cow. 
     \par \textbf{Rules}: if something sees the squirrel and the squirrel eats the cow then the cow is round. if something is green then it eats the tiger. if the squirrel is round then the squirrel visits the cow. if something eats the cow then it sees the squirrel. if something sees the tiger and the tiger visits the squirrel then it is nice. if something is round then it eats the cow. if something is kind then it eats the cow. if the tiger visits the cow then the cow sees the squirrel. if something sees the cow then the cow eats the tiger.
      \par
    \textbf{Question:} The cow does not see the squirrel.  
    & \textbf{Step1}: The squirrel is round. \par
      \textbf{Step2}: something is round then it eats the cow. \par 
     \textbf{Step3}: The squirrel eats the cow. \par
     \textbf{Step4}: If something sees the squirrel and the squirrel eats the cow then the cow is round. \par
     \textbf{Step5}: The cow is round.  \par
     \textbf{Step6}: If something is round then it eats the cow. \par
     \textbf{Step7}: The cow eats the cow. \par
     \textbf{Step8}: if something eats the cow then it sees the squirrel.\par
     \textbf{Step9}: the cow sees the squirrel. \par \textbf{Answer}: True \\
     \midrule
   \textbf{MATH} & \textbf{Context}: Tom has a red marble, a green marble,
a blue marble, and three identical yellow marbles. \par
\textbf{Question}: How many different groups of two marbles can
Tom choose?  &   \textbf{Step1}: There are two cases here: \par \textbf{Step2}: either Tom chooses two yellow marbles (1 result), or he
chooses two marbles of different colors ($\binom{4}{2}$=6 results.). \par\textbf{Step3}: The total number of distinct pairs of marbles Tom can choose is 1 + 6 = 7. \par \textbf{Answer}: 7 \\
\midrule
\textbf{ASDIV} & \textbf{Context}: A sandwich is priced at \$0.75. 
    A cup of pudding is priced at \$0.25. Tim bought 2 sandwiches and 4 cups of pudding. \par  \textbf{Question}: How much money should Tim pay? & 0.75 x 2 + 0.25 x 4 = 2.5 \par \textbf{Answer}: 2.5 \\
    \midrule
       \textbf{AQUA} & \textbf{Context}: The entrance fee for a fair is \$5 for persons under the age of 18 and 20\% more for persons older. Each ride at the fair costs \$0.50. If Joe goes with her 6 years old twin brothers, and they each took 3 rides in total. \par \textbf{Question:} How much money does
    Joe end up spending at the fair?  &  
    \textbf{Step1}: Total entrance fee is (2*\$5)+(1.20*5) = \$16 \par 
    \textbf{Step2}: Total rides fee is ( 0.50 * 3 ) * 3 = \$4.50 \par
    \textbf{Step3}: Total money spent is \$20.50 \par \textbf{Answer}: 20.5 \\\midrule
       \textbf{EQASC} & \textbf{Question}: Where is water likely to form beads? & 
    \textbf{Step1}: Beads of water are formed by water vapor condensing \par
     \textbf{Step2}: Moisture builds up in condenses air and the wherever the surfaces are cold. \par \textbf{Answer}: Water beads form on cold surfaces. \\
\midrule
\textbf{StrategyQA} &  \textbf{Question}: Are more people today related to Genghis Khan than Julius Caesar? & 
    \textbf{Step1}: Julius Caesar had three children. \par 
    \textbf{Step2}: Genghis Khan had sixteen children. \par
    \textbf{Step3}: Modern geneticists have determined that out of every 200 men today has DNA that can be traced to Genghis Khan. \par  \textbf{Answer}: True \\
\bottomrule
    \end{tabular}%
    }
  \vspace{-10pt}
\end{table}

\textbf{EntailmentBank (EntBank)}~\citep{entalmentbank2021} is a complex question answering dataset which contains multi-step entailment trees, namely a tree of multi-premise entailment steps from facts that are known, through intermediate conclusions to hypothesis of interest (which in this case the question and answer).

\textbf{ProofWriter}~\citep{tafjord-etal-2021-proofwriter} is a question answering dataset for logical reasoning. It contains 500k questions, answers and proofs over natural-language rulebases. This dataset is mostly used to emulate reasoning over rules expressed in language, including proof generation. The datasets proofs include intermediate conclusions. In our experiments, we used \textit{depth-0}, \textit{depth-1}, \textit{depth-2}, \textit{depth-3}, and \textit{depth-5} \textit{OWA} sets.

\textbf{MATH}~\citep{hendrycksmath2021} is a dataset of 12,500 problems from high school math competitions. Given a math problem such as in Table~\ref{alldatasetssamples-diaognostic} models generate a sequence, such as $\frac{2}{3}$, that encodes
the final answer.

\textbf{ASDIV}~\citep{miao-etal-2020-diverse} (Academia Sinica Diverse MWP Dataset) is a dataset of 2,305 questions on diverse math word problem solving. It includes a diverse operations such as basic arithmetic or aggregative operations (e.g., comparisons, set-operations). 

\textbf{AQUA}~\citep{liang-etal-2018-meaning} is a dataset of 100,000 algebraic word problems with step-wise solutions as shown below. 
In the original dataset each question is decomposed in four parts, two inputs and two
outputs: the description of the problem and a question, and the possible (multiple choice) answer options, one being the correct one. In this work we only used the context and question, the step-wise solution and the correct answer to construct our diagnostic dataset.

\textbf{EQASC}~\citep{aggarwaletal2021ecqa} is a multi-hop question answering dataset with 98K explanation annotations for multi-step factual reasoning. Each instance in the dataset comes with a question, multiple answer choices, explanation of each answer choice and a free flow explanation of the whole context. In our experiments we used the correct answer's explanation to construct our diagnostic datasets. 

\textbf{StrategyQA}~\citep{geva2021did} is another multi-step question answering (QA) dataset, that covers a diverse set
of reasoning skills. StrategyQA consists of 2,780 questions, annotated with their decomposition and per-step evidence. 

\subsection{Human Judged Dataset Construction} 
\label{app:humanjudgeddatasets}
In the following we present details of each human judged datasets used in our work. Table~\ref{tab:dataset-appendix} lists each dataset and illustrates how each dataset is used in our experiments. Specifically, all six datasets are used for evaluations in the experiments results and model finetuning, and one dataset was used for finetuning only. The dataset details are explained below.  

To construct these datasets, we first sample instances from each dataset (see the number of instances sampled in Table~\ref{tab:dataset-appendix}). We use GPT-3 with few-shot in-context examples and a prompt to generate step-by-step reasoning (e.g., "\textit{explain step-by-step}") for each sampled instance (see in-context examples and prompts in App.~\ref{app:intro}). Then, using our taxonomy we constructed a list of evaluation perspectives to label the model generated step-by-step reasoning step of each of these datasets.  We explain the details of the perspectives used to label human judged datasets in $\S$~\ref{experimentsetup} and App.~\ref{app:annotation}. All datasets with examples are summarised in in Table~\ref{alldatasetssamples-human}. In the following we present details of each human judged datasets.  

\begin{table}[h!]
\caption{\footnotesize We show instances from five of the \textbf{Human Judged Datasets} used in our work. Only GSM8K and ESNLI include human labeled explanations.}
\label{alldatasetssamples-human}
\begin{center}
    \scalebox{0.75}{
  \begin{tabular}{@{}lp{290pt}p{200pt}@{}}
    \toprule
    \textbf{Dataset} & \textbf{Instance} & \textbf{Reference Answer \&  Reference Step-by-Step Solution}  \\
    \midrule
    \textbf{GSM8K} & \textbf{Question}: Tina buys 3 12-packs of soda for a party. Including Tina, 6 people are at the party. Half of the people at the party have 3 sodas each, 2 of the people have 4, and 1 person has 5. How many sodas are left over when the party is over? \par & 
    \textbf{Answer}: 11 \par
    \textbf{Step1}: Tine buys 3 12-packs of soda for 3*12=36 sodas \par 
    \textbf{Step2}: 6 people attend the party, so half of them is 6/2= 3 people \par
    \textbf{Step3}: Each of those people drinks 3 sodas, so they drink 3*3=9 sodas. \par 
    \textbf{Step4}: Two people drink 4 sodas, which means they drink 2*4=8 sodas. \par
    \textbf{Step5}: With 1 person drinking 5, that brings the total drank to 5+9+8+3=25 sodas \par
    \textbf{Step6}: As Tina started off with 36 sodas, that means there are 36-25=11 sodas left.\\\midrule
    \textbf{CosmosQA} &     \textbf{Context}: A woman had topped herself by jumping off the roof of
the hospital she had just recently been admitted to. She was
there because the first or perhaps latest suicide attempt was
unsuccessful. She put her clothes on, folded the hospital gown
and made the bed. She walked through the unit unimpeded and
took the elevator to the top floor
\par
\textbf{Question}: What would have happened to the woman if the staff at
the hospital were doing their job properly?
    & \textbf{Answer}: The woman would have been stopped before she left to
take the elevator to the top floor and she would have lived. \\ \midrule
\textbf{DROP} & \textbf{Context}: Denver would retake the lead with kicker Matt Prater
nailing a 43-yard field goal, yet Carolina answered as
kicker John Kasay ties the game with a 39-yard field
goal. . . . Carolina closed out the half with Kasay nailing a 44-yard field goal. . . . In the fourth quarter, Carolina sealed the win with Kasay’s 42-yard field goal. \par \textbf{Question}: Which kicker kicked
the most field goals? & \textbf{Answer}: John
Kasay \\\hline
\textbf{ESNLI} & \textbf{Premise}: A child in a yellow plastic safety swing is laughing as a dark-haired woman
in pink and coral pants stands behind her. \par
\textbf{Hypothesis}: A young mother is playing with her daughter in a swing. &
\textbf{Answer}: neutral \par
\textbf{Explanation}: Child does not imply daughter and woman does not imply mother. \\ \midrule
\textbf{SemEVAL} & \textbf{Context}: Now I am going to set the dining table up for dinner. First I 
             put away all the stuff that is not supposed to be on the table. 
             Next I clean the table with a tissue paper. Then I arrange some 
             of the decorations on the table. After that I put down the 
             plates and glasses.lastly in the remaining spaces on the table I 
             put down what we are going to eat but I dont put down dessert 
             yet. There is one more thing I do before I am finished setting 
             up the dining table. I call my family down for dinner. I swept a 
             drop of sweat of my chin and forehead. It was hard hard work but 
             still it was so much fun. Oh no my family is done with dinner 
             and now I have to bring them dessert which is fruit I made 
             myself. \par \textbf{Question}: When did they clean the dining table? 
             \par \textbf{Hypothesis}: After it was set. &  \textbf{Answer}: No
             \\
\bottomrule
    \end{tabular}%
    }
\end{center}
\end{table}

\textbf{DROP}~\citep{dua-etal-2019-drop}, Discrete
Reasoning Over the content of Paragraphs, is a dataset of 96K of instances with context and a question. To solve the tasks, a system must resolve
references in the context that match with the question, and perform discrete operations
over them (such as addition, counting, or sorting).  These operations require
comprehensive understanding of the content of the input
context. 

\textbf{GSM8K}~\citep{cobbe2021gsm8k} is a dataset of 8.5K linguistically diverse grade school math word problems. On this dataset, even the largest transformer models fail to achieve high test performance, despite the conceptual simplicity of this problem distribution.

\textbf{CosmosQA}~\citep{huang-etal-2019-cosmos} is a dataset of 35K problems that
require commonsense-based reading comprehension, formulated as multiple-choice questions. 
The questions
focus on reading between the lines over a diverse collection of people’s everyday narratives, asking such questions as "\textit{what might be the possible reason of ...?}", or "\textit{what would have happened if ...?}". The dataset does not introduce step-by-step reasoning output, and contains multiple choice answers.

\textbf{ESNLI}~\citep{esnli} is the extended version of the Stanford Natural Language Inference
corpus~\citep{bowman-etal-2015-large} of 570K labeled sentence pairs with entailment or contradiction labels. ESNLI includes human labeled explanations of the entailment decision. 

\textbf{SemEVAL}~\citep{Ostermann2018SemEval2018T1} is a dataset on machine comprehension
using commonsense knowledge. It contains questions that require commonsense
knowledge for finding the correct answer. 

\subsection{Synthetic Diagnostics Dataset Generation with Perturbation Rules}
\label{app:datasetconstruction}
To construct the diagnostics datasets we apply synthetic perturbations on half of the chains from six datasets (for details see App.~\ref{app:diagnosticdatasets} and the summary Table~\ref{tab:dataset-appendix}). Also, in Table~\ref{tab:perturbations} we illustrate these synthetic perturbations applied on reasoning steps $\{r_i\}$ of gold reference chains of all the datasets. In there, $\bm{g^*}$ indicates a grammar error, which includes changing verb tense, dropping verb, or random word swap. $\bm{s^*}$ represents change the semantics of one step in the chain by replacing named entities. To simulate extrinsic hallucinations, we use random steps from other chains within the same dataset.

\begin{table}[t!]
        \caption{\label{tab:perturbations} \footnotesize Synthetic perturbations and corresponding error types of steps $\{r_i\}$ in reference chains used when constructing diagnostics datasets. $\bm{g^*}(\cdot)$ represents grammar error, $\bm{s^*}(\cdot)$ represents semantic change. 
        }    
    \scalebox{0.85}
	    {
    \begin{tabular}{llcl}
    \toprule
    \textbf{Perturbation Type} & \textbf{Error Type} & \textbf{Reference Reasoning Steps} & \textbf{Hypothesis Reasoning Steps}  \\
    \midrule
     
    {Repeat a step} & {Repetition} & \texttt{$[r_1, r_2, r_3]$} & \texttt{$[r_1, r_2, r_2, r_3]$} \\ 
    {Remove a step} & {Missing step} & \texttt{$[r_1, r_2, r_3]$} & \texttt{$[r_2, r_3]$} \\
    {Shuffle steps} & {Self-coherency} & \texttt{$[r_1, r_2, r_3]$} & \texttt{$[r_3, r_1, r_2]$} \\
    {Swap a step} & {Self-coherency} & \texttt{$[r_1, r_2, r_3]$} & \texttt{$[r_2, r_1, r_3]$} \\ 
    {Negate a step} & {Factuality} & \texttt{$[r_1, r_2, r_3]$} & \texttt{$[r_1, \neg r_2, r_3]$}
    \\
    {Hallucination} & {Hallucination} & \texttt{$[r_1, r_2, r_3]$} & \texttt{$[r_1, r_2, r_3, r_4]$}
    \\
    {Grammar error} & {Grammatical} & \texttt{$[r_1, r_2, r_3]$} & \texttt{$[r_1, r_2, \bm{g^*}(r_3)]$}
    \\
    {Semantic change} & {Factuality} & \texttt{$[r_1, r_2, r_3]$} & \texttt{$[r_1, \bm{s^*}(r_2), r_3]$}
    \\

\bottomrule
     \vspace{5pt}
 \end{tabular}%
 }
 
\end{table}

To construct diagnostic data from math datasets, we introduce four additional perturbations to simulate step-wise explanation errors that might arise in arithmetic reasoning task (\textit{Arithmetic error}), general knowledge about relationships and equation construction (\textit{Common sense error}), and misinformation about object/subject characteristics (\textit{Factuality} or \textit{Hallucination}): 
\begin{itemize} \itemsep-0.3em
    \item \textbf{Shuffle numbers}: randomly shuffles all numbers in the chain,
    \item \textbf{Shuffle operations}: randomly shuffles all math operations in the chain,
    \item \textbf{Random number}: randomly replaces one number in the chain,
    \item \textbf{Random operation}: randomly replaces one math operation in the chain.
\end{itemize}

\clearpage
\section{Human Annotations (Cont. from $\S$~\ref{experimentsetup})}
\label{app:annotation}
\begin{figure}[h!]
\caption{\footnotesize Screenshot of expert annotation user interface, showing the context for the initial question as well as the questions regarding the generated response.}
\label{fig:annotation_ui_whole}
\centering
\includegraphics[origin=c,width=0.95\linewidth]{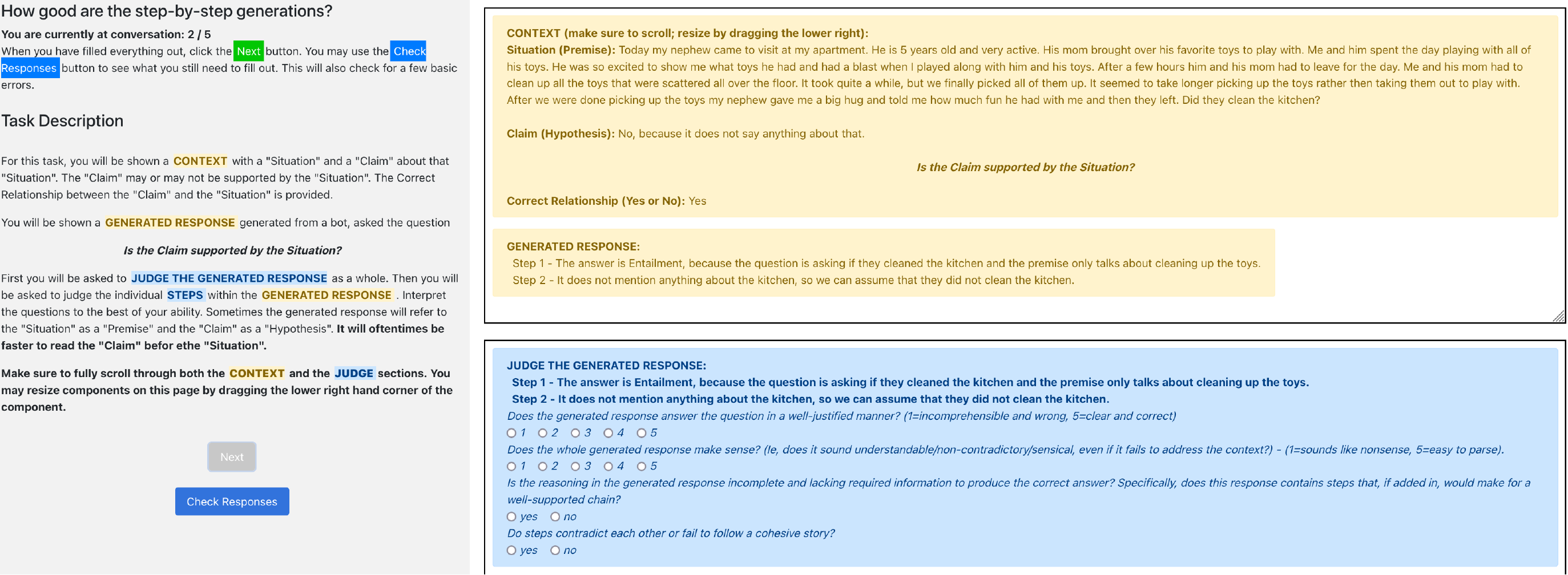}
\end{figure}

\begin{figure}[h!]
\caption{\footnotesize Screenshot of expert annotation user interface, showing questions asked for each step, using the question in Fig \ref{fig:annotation_ui_whole}. The questions are asked of every step generated by the model, with steps separated by sentence-ending periods.}
\label{fig:annotation_ui_step}
\centering
\includegraphics[origin=c,width=0.95\linewidth]{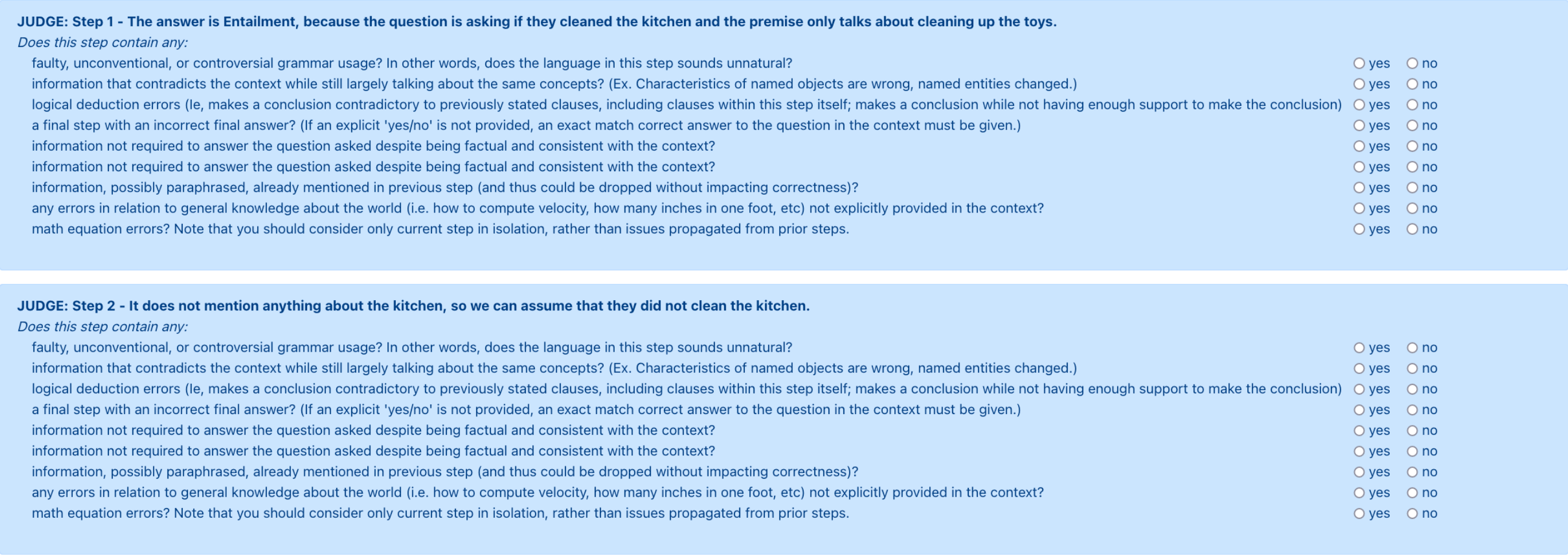}
\end{figure}

To construct \textbf{Human Judged Datasets}, we perform human annotations on five datasets which we summarize in Table~\ref{tab:dataset-appendix} (Type='Human judged'). These datasets do not include explanations (except GSM8K and ESNLI), so we construct model generated reasoning steps and label them with reasoning errors. We explain our generation process in $\S$\ref{experimentsetup} and App.~\ref{app:humanjudgeddatasets}. 
We used five expert human annotators to collect reasoning error labels on five datasets. 
We asked human evaluators to directly rate
the generated reasoning errors on overall chain level using a Likert scale from 1 to 5. We also asked them
to mark whether each error type proposed in our error taxonomy ($\S$\ref{taxonomy}) appeared in each 
step in step-level evaluations. In Fig.~\ref{fig:annotation_ui_whole} and Fig.~\ref{fig:annotation_ui_step} we illustrate the UI used to collect the data. Table~\ref{tab:appx-perspectives} summarizes questions that experts were asked. Table~\ref{tab:human-eval-error-stats} reports the distribution of errors for each dataset.
In general, we found that it was hard to get anonymous crowd workers to annotate our data accurately even when we paid averages of upwards of \$30 an hour, hence relying on expert annotators. For the annotation sessions reported in the text of the paper, we find that it takes an average of 754 seconds for expert annotators to complete a session of at most 5 examples, or slightly over 2-and-a-half minutes per example. This highlights the difficulty of obtaining high-quality annotations on these cognitive challenging tasks.

\vspace{5pt}
\vspace{15pt}

\begin{table*}[t]
  \centering
  \footnotesize
      \caption{\label{tab:human-eval-perspectives} \footnotesize Evaluation perspectives used to \textbf{Human Judged} the datasets.  The perspectives, which we used to ask humans to label, align with our taxonomy of reasoning errors. (Continued from $\S$~\ref{datasets}) 
      }
      \label{tab:appx-perspectives}
      \scalebox{0.9}
	    {
     \renewcommand{\arraystretch}{1.3}
\begin{tabular}{@{}llp{80pt}lp{250pt}@{}}
\toprule
\textbf{Level} & & \textbf{Evaluation} & \textbf{Label} 
& \textbf{Details}  \\
& & \textbf{Perspective} &  \\
\midrule
Overall & QUAL & Overall quality & [1-5] 
& Does the generated response answer the question in a well-justified manner? (1=incomprehensible and wrong, 5=clear and correct) \\

Overall & COH & Coherency & [1-5] 
& Does the whole generated response make sense? (Ie, does it sound understandable/non-contradictory/sensical, even if it fails to address the context?) - (1=sounds like nonsense, 5=easy to parse). \\
\midrule
Step & MISS & Missing Step & Y/N 
&  Is the reasoning in the generated response incomplete and lacking required information to produce the correct answer? Specifically, does this response contains steps that, if added in, would make for a well-supported chain?  \\


Step & GRAM& Grammar & Y/N  
& Does this step contain faulty, unconventional, or controversial grammar usage? In other words, does the language in this step sounds unnatural?  \\ 

Step & FACT & Factuality  & Y/N 
& Does this step contain information that contradicts the context while still largely talking about the same concepts? (Ex. Characteristics of named objects are wrong, named entities changed.) \\

Step & LOGIC & Coherency and Logic & Y/N  
& Does this step any logical deduction errors (Ie, makes a conclusion contradictory to previously stated clauses, including clauses within this step itself; makes a conclusion while not having enough support to make the conclusion) \\ 


Step & HALL & Hallucination & Y/N  
&  Does this step contain information not provided in the problem statement that is irrelevant or wrong? \\ 

Step & RED & Redundancy  & Y/N  
&  Does this step contain information not required to answer the question asked despite being factual and consistent with the context? \\ 

Step & REP & Repetition  & Y/N  
&  Does this step contain any information, possibly paraphrased, already mentioned in previous step (and thus could be dropped without impacting correctness)? \\ 

Step & COMMON &  Commonsense  & Y/N  
&  Does this step contain any errors in relation to general knowledge about the world (i.e. how to compute velocity, how many inches in one foot, etc) not explicitly provided in the context? \\ 

Step & MATH & Arithmetic  & Y/N  
&  Does this step contain math equation errors? Note that you should consider only current step in isolation, rather than issues propagated from prior steps. \\ 

\bottomrule
\end{tabular}
}
\end{table*}
\begin{table*}[t]
\vspace{-10pt}
  \centering
  \footnotesize
      \caption{\label{tab:human-eval-error-stats} \footnotesize Statistics of types of errors in \textbf{Human Judged} datasets. Each column reports the number of examples where the specified error type exists. (Continue from $\S$~\ref{datasets})}
\begin{tabular}{@{}lcccccc}
\toprule
  Error Type
  & \multicolumn{1}{@{}c}{\textbf{DROP}}
  & \multicolumn{1}{@{}c}{\textbf{GSM8K}}
  & \multicolumn{1}{@{}c}{\textbf{ESNLI}}
  & \multicolumn{1}{@{}c}{\textbf{COSMOS}}
  & \multicolumn{1}{@{}c}{\textbf{SemEVAL}}  \\
\midrule
Grammar & $8$ & $4$ & $5$ & $8$ & $6$\\ 
Factuality & $19$ & $56$ & $15$ & $44$ & $31$\\
Hallucination & $4$ & $8$ & $4$ & $9$ & $2$\\ 
Redundancy & $25$ & $13$ & $14$ & $15$ & $19$\\ 
Repetition & $2$ & $2$ & $0$ & $3$ & $3$\\
Missing Step & $109$ & $81$ & $40$ & $99$ & $67$\\
Coherency & $20$ & $57$ & $17$ & $48$ & $17$\\ 
Commonsense & $3$ & $58$ & $5$ & $18$ & $1$\\
Arithmetic & $2$ & $7$ & $1$ & $0$ & $0$\\ 
\bottomrule
\end{tabular}
\end{table*}

\clearpage
\section{Sentence Embedding Model Training (Cont. from $\S$\ref{experiments})}
\label{app:finetuning}
\textbf{Model training.} We use the train portions of the perturbed diagnostics datasets to finetune the SimCSE embeddings model (explained in \S~\ref{training}) and validation portions to select the best embedding model. The test portions are used to evaluate our metrics against baseline metrics. We randomly select 500,000 samples with replacement from each dataset to create uniform representation and reduce bias.

The hyperparameters used to finetune SimCSE model are described in Table~\ref{tab:hyperparam-roscoe}. We use NVIDIA Tesla V100 Volta GPU instances with 32GB Graphics Card. We perform hyperparameter search, varying batch size in $\{32, 64, 256, 512, 1024, 2048\}$, learning rate in $\{5e\textnormal{-}06, 1e\textnormal{-}05, 5e\textnormal{-}05, 1e\textnormal{-}04\}$, and max sequence length in $\{64, 128, 512\}$. Not all combinations of batch size and max sequence length were explored due to memory limitations. 
\vspace{0.5cm}
\begin{table*}[h!]
\vspace{-10pt}
  \centering
  \footnotesize
\caption{\label{tab:hyperparam-roscoe} Hyperparameters used to fine-tune SimCSE model on perturbed datasets.}
\begin{tabular}{lc}
\toprule
Parameter & Value  \\
\midrule
Batch size & 64 \\
Max sequence length & 512 \\
Training epochs & 5 \\
Learning rate & 5e-6 \\
Temperature & 0.05 \\
\bottomrule
\end{tabular}
\end{table*}

\textbf{Validation.} We replace original validation procedure on semantic textual similarity tasks with similarity-based validation on perturbed reasoning chains. In particular, during training, we select best checkpoint that maximizes cosine similarity between positive and minimizes cosine similarity between hard-negative pairs within the batch of size $B$ as the following:
\begin{align}
    \frac{\textstyle\sum_{i=1}^{N}\left[\cos(s_i, r_i) - \cos(s_i, h_i)\right]}{2*B}
\end{align}
Model is evaluated every 100 steps on the development dataset and the best checkpoint is applied at the inference. Other parameters not described in this section are kept as in the original SimCSE model used for initialization.

\textbf{Inference}. We compare \ourmodel~scores calculated against three embeddings: finetuned SimCSE model, \textit{sup-simcse-roberta-base} SimCSE model, and \textit{all-mpnet-base-v2} sentence embedding model~\citep{reimers-2019-sentence-bert}.
During inference, we set the random seed to 42. Without this, the embedding-based scores naturally varied by about 0.01.

\newpage
\section{Additional Experimental Results  (Cont. from $\S$\ref{experiments})}
\label{app:additionalresults}
\subsection{Controlled Experiments with Diagnostics Datasets}
\label{app:diagnostics-experiments}
In this section, we presented Somers' $D$ correlation of all metrics on all Diagnostics datasets. Table~\ref{tab:corr-synth-unsup-max} summarizes the evaluations when investigated reference-free. One of the characteristics of our \ourmodel~metrics is that, they can provide judgement of the model generated reasoning steps with and without the human reference reasoning chains. In the experiments section in \S\ref{experiments}, we discussed the results of our unsupervised scores in comparison to baseline scores when measured reference-free. In Table~\ref{tab:corr-synth-sup-max}, we summarize the correlation analysis on \ourmodel~metrics in comparison to baselines on diagnostic datasets when reference is present for evaluation. Specifically, each score is measured between the human provided reasoning steps (\textit{reference}) and the model generated reasoning steps (\textit{hypothesis}). We also display fine-grained meta-evaluations of all metrics on each diagnostics dataset in separate tables. Specifically,  
Tables~\ref{tab:corr-synth-unsup-perturb-eqasc},~\ref{tab:corr-synth-sup-perturb-eqasc} for EQASC, Tables~\ref{tab:corr-synth-unsup-perturb-entailment_bank},~\ref{tab:corr-synth-sup-perturb-entailment_bank} for EntailmentBank,
Tables~\ref{tab:corr-synth-unsup-perturb-math_dataset},~\ref{tab:corr-synth-sup-perturb-math_dataset} for MATH,
Tables~\ref{tab:corr-synth-unsup-perturb-proofwriter},~\ref{tab:corr-synth-sup-perturb-proofwriter} for ProofWriter,
Tables~\ref{tab:corr-synth-unsup-perturb-asdiv},~\ref{tab:corr-synth-sup-perturb-asdiv} for ASDIV, and
Tables~\ref{tab:corr-synth-unsup-perturb-aqua},~\ref{tab:corr-synth-sup-perturb-aqua} for AQUA.

\begin{table*}[b]
  \centering
  \footnotesize
  \setlength\tabcolsep{3.4pt}
  \renewcommand{\arraystretch}{1}
          \caption{\footnotesize Somers' $D$ correlation of \textbf{all} metrics on six \textbf{Diagnostics} datasets. All metrics are measured \textbf{reference-free} on $(\bm{s},\bm{h})$. The highest correlation overall for each dataset is in \textbf{bold}. The second best models are \underline{underlined}. Correlations that are not significant (\textit{p-value} $>= 0.05$) are omitted when aggregating, and "-" denotes an absence of any significant correlations. Note that ASDIV is a 1-step equation dataset, so there are no repetition and self-consistency scores as there are no steps to compare. 
          (Continued from $\S$\ref{diagnostics-experiments}, more details in App.~\ref{app:diagnostics-experiments}.).}
    \scalebox{0.9}
    {
    \begin{tabular}{lcccccc}
    \toprule
          \textbf{Metric} & \multicolumn{1}{c}{\textbf{EntBank}} & \multicolumn{1}{c}{\textbf{Math}} & \multicolumn{1}{c}{\textbf{\textbf{AQUA}}} & \multicolumn{1}{c}{\textbf{ProofWriter}} & \multicolumn{1}{c}{\textbf{EQASC}} & \multicolumn{1}{c}{\textbf{ASDIV}} \\
           \midrule
    ROUGE-1  & 0.410 & 0.176 & 0.257 & 0.095 & 0.342 & 0.305\\
    ROUGE-2  & 0.391 & 0.151 & 0.206 & 0.090 & 0.217 & - \\
    ROUGE-L  & 0.365 & 0.156 & 0.264 & 0.106 & 0.315 & 0.269\\
    BLEURT  & 0.257 & 0.148 & 0.252 & 0.024 & 0.447 & - \\
    BERTScore  & 0.380 & 0.124 & 0.220 & 0.117 & 0.462 & 0.322\\
    \bartscore  & 0.358 & 0.185 & 0.317 & 0.081 & 0.415 & - \\
    \bartscoreParaCnn  & 0.315 & 0.164 & 0.251 & 0.054 & 0.297 & - \\
\bottomrule
    \end{tabular}%
    }
  \label{tab:corr-synth-unsup-max}%
\end{table*}%
\begin{table*}[h]
  \centering
  \footnotesize
  \setlength\tabcolsep{3.4pt}
  \renewcommand{\arraystretch}{1}
          \caption{\footnotesize Somers' $D$ correlation of all \textbf{reference-based} metrics on six \textbf{Diagnostics} datasets. Metrics are measured using \textbf{reference} generations on $(\bm{r},\bm{h})$. The highest correlation overall for each dataset is in \textbf{bold}. The second best models are \underline{underlined}. 
          (Continued from $\S$\ref{diagnostics-experiments}, more details in App.~\ref{app:diagnostics-experiments}.)}
    \scalebox{0.9}
    {
    \begin{tabular}{lcccccc}
    \toprule
          \textbf{Metric} & \multicolumn{1}{c}{\textbf{EntBank}} & \multicolumn{1}{c}{\textbf{Math}} & \multicolumn{1}{c}{\textbf{\textbf{AQUA}}} & \multicolumn{1}{c}{\textbf{ProofWriter}} & \multicolumn{1}{c}{\textbf{EQASC}} & \multicolumn{1}{c}{\textbf{ASDIV}} \\
  \midrule
     ROUGE-1  & \textbf{1.000} & \textbf{1.000} & \textbf{1.000} & \textbf{1.000} & \textbf{1.000} & \textbf{1.000}\\
    ROUGE-2  & \textbf{1.000} & \textbf{1.000} & \textbf{1.000} & \textbf{1.000} & \textbf{1.000} & \textbf{1.000}\\
    ROUGE-L  & \textbf{1.000} & \textbf{1.000} & \textbf{1.000} & \textbf{1.000} & \textbf{1.000} & \textbf{1.000}\\
    BLEURT  & 0.821 & 0.773 & 0.946 & 0.829 & \textbf{1.000} & 0.93\\
    BERTScore  & \textbf{1.000} & \textbf{1.000} & \textbf{1.000} & \textbf{1.000} & \textbf{1.000} & \textbf{1.000}\\
    \bartscore  & 0.983 & 0.989 & \underline{0.984} & 0.986 & \textbf{1.000} & \underline{0.954}\\
    \bartscoreParaCnn  & 0.988 & 0.963 & \textbf{1.000} & 0.996 & \textbf{1.000} & \textbf{1.000}\\
\bottomrule
    \end{tabular}%
    }
  \label{tab:corr-synth-sup-max}%
\end{table*}%
\input{tables/table-appx-corr-syn-unsup-perturb-eqasc.tex}
\input{tables/table-appx-corr-syn-unsup-perturb-entailment_bank.tex}
\input{tables/table-appx-corr-syn-unsup-perturb-math_dataset}
\input{tables/table-appx-corr-syn-unsup-perturb-proofwriter.tex}
\input{tables/table-appx-corr-syn-unsup-perturb-asdiv.tex}
\input{tables/table-appx-corr-syn-unsup-perturb-aqua.tex}

\input{tables/table-appx-corr-syn-sup-perturb-eqasc.tex}
\input{tables/table-appx-corr-syn-sup-perturb-entailment_bank.tex}
\input{tables/table-appx-corr-syn-sup-perturb-math_dataset}
\input{tables/table-appx-corr-syn-sup-perturb-proofwriter.tex}
\input{tables/table-appx-corr-syn-sup-perturb-asdiv.tex}
\input{tables/table-appx-corr-syn-sup-perturb-aqua.tex}

\begin{figure}[h!]       
    \mbox{\includegraphics[height=.4\textheight]{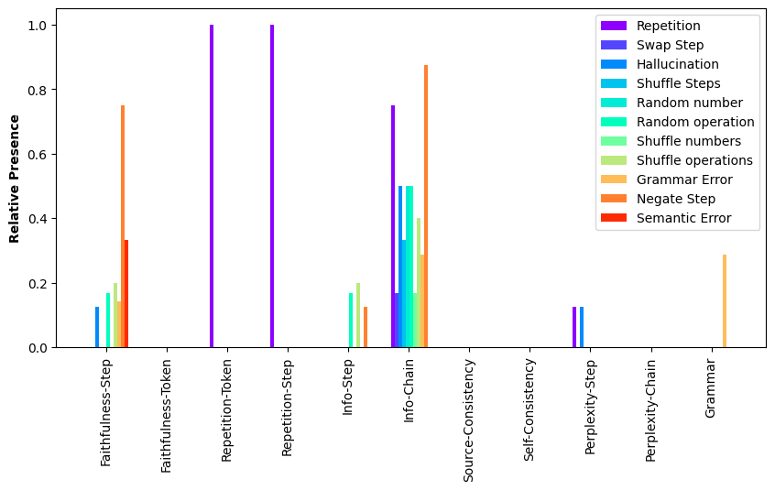}}
    \hspace{-8px}
    \caption{Relative presence of the strong score-perturbation correlation, measured as the number of datasets where for each score-perturbation pair Somers' $D$ correlation value is in the $90^{th}$ percentile, normalized by the total number of datasets where this type of perturbation occurs. Statistics collected over \ourmodel~reference-free scores with finetuned SimCSE embeddings. (Continued from $\S$\ref{analysis})}
    \label{fig:score-perturb}
\end{figure}

To understand if designed reference-free scores capture targeted error types we analyze perturbation-level correlations summarized in Fig.~\ref{fig:score-perturb}. 
Out of the all considered scores, \textit{Info-Chain} is able to cover 10 out of 12 of errors, except \textit{Remove Step} and \textit{Semantic error} perturbations. In general we can note that \ourmodel~fails to consistently identify \textit{missing step} error type represented by \textit{Remove Step} perturbation across different datasets, while other synthesized error types are covered by at least one score type.

Reference-based scores are covering all synthetic errors, with Semantic Coverage Chain showing strong correlations with all types of perturbations (Table~\ref{tab:corr-synth-sup-max}). We also note that along with \ourmodel~scores, the highest correlation among all reference-based scores belong to ROUGE and BERT scores (Tables~\ref{tab:corr-synth-sup-perturb-eqasc}-\ref{tab:corr-synth-sup-perturb-aqua}). ROUGE scores consistently  outperform on \textit{Repetition}, \textit{Hallucination}, \textit{Remove Step}, \textit{Shuffle Steps}, \textit{Swap Steps}, \textit{Negate Step}, and \textit{Semantic} perturbations, while under performing on \textit{Random operation}, and \textit{Shuffle operations}. We attribute this to the fact that ROUGE is an n-gram based score, so it is better in catching errors were wording has significantly changed, while failing to catch small changes within steps.

It is worth noting that some scores, especially those among reference-based evaluations, get the highest possible Somers' \textit{D} correlation scores of $1.0$. What it means is that in some scenarios, there is a perfect correlation between the metric and the error type. In other words, for this metric we can find a threshold such generated chains that have scores greater than the threshold do not have errors of the given type, and in all generated chains with scores less than the threshold have that error. It is especially evident on referenced-based metrics that directly compare the reference solution and hypothesis. In this scenario, we build correlation for two groups: 1) non-perturbed hypothesis: the score is calculated by comparing embedding similarities of the reference with itself, and we expect to get high scores, 2) perturbed hypothesis: comparing reference with its perturbed version, where the scores should be lower. In some cases, we are able to perfectly separate perturbed and non-perturbed chains based on the corresponding metric values by selecting a threshold, in other cases we cannot due to a number of false-negatives (i.e., a chain gets a high score, although the error is present).
As an example, consider the \textit{Semantic Coverage-Chain} metric calculated on EQASC dataset using \textit{all-mpnet-base-v2} sentence embeddings, and \textit{Hallucination} perturbation (Table~\ref{tab:corr-synth-sup-perturb-eqasc}). Here the Somers' \textit{D} correlation score is $1.0$. \textit{Semantic Coverage-Chain} is calculated as a normalized cosine distance between the chain embedding of the reference solution $\vr$, and the chain embedding of the hypothesis $\vh$ : $[1 + \cos(\vr, \vh)] / 2$. Recall that in our setup, half of the hypothesis chains are perturbed reference chains, and another half is the same as the reference.  While \textit{Hallucination} perturbation is an insertion of a random step from a dataset, it is hard to predict how if will affect the embedding of the chain as a whole, but on the unperturbed chains, where $\vh==\vr$, the \textit{Semantic Coverage-Chain} should be: $[1 + \cos(\vr, \vr)] / 2 = 1.0$. Further review confirmed that in this dataset there are no false-positive instances, i.e., all chains with perturbations had \textit{Semantic Coverage-Chain} score less than $1.0$. That means, we can always identify if the chain contains a \textit{Hallucination} error or not, by comparing \textit{Semantic Coverage-Chain} value with $1.0$ (threshold value), which is reflected in perfect Somers' \textit{D} score.

\begin{figure}[h!]   
\vspace{0.9em}
    \mbox{\includegraphics[height=.2\textheight]{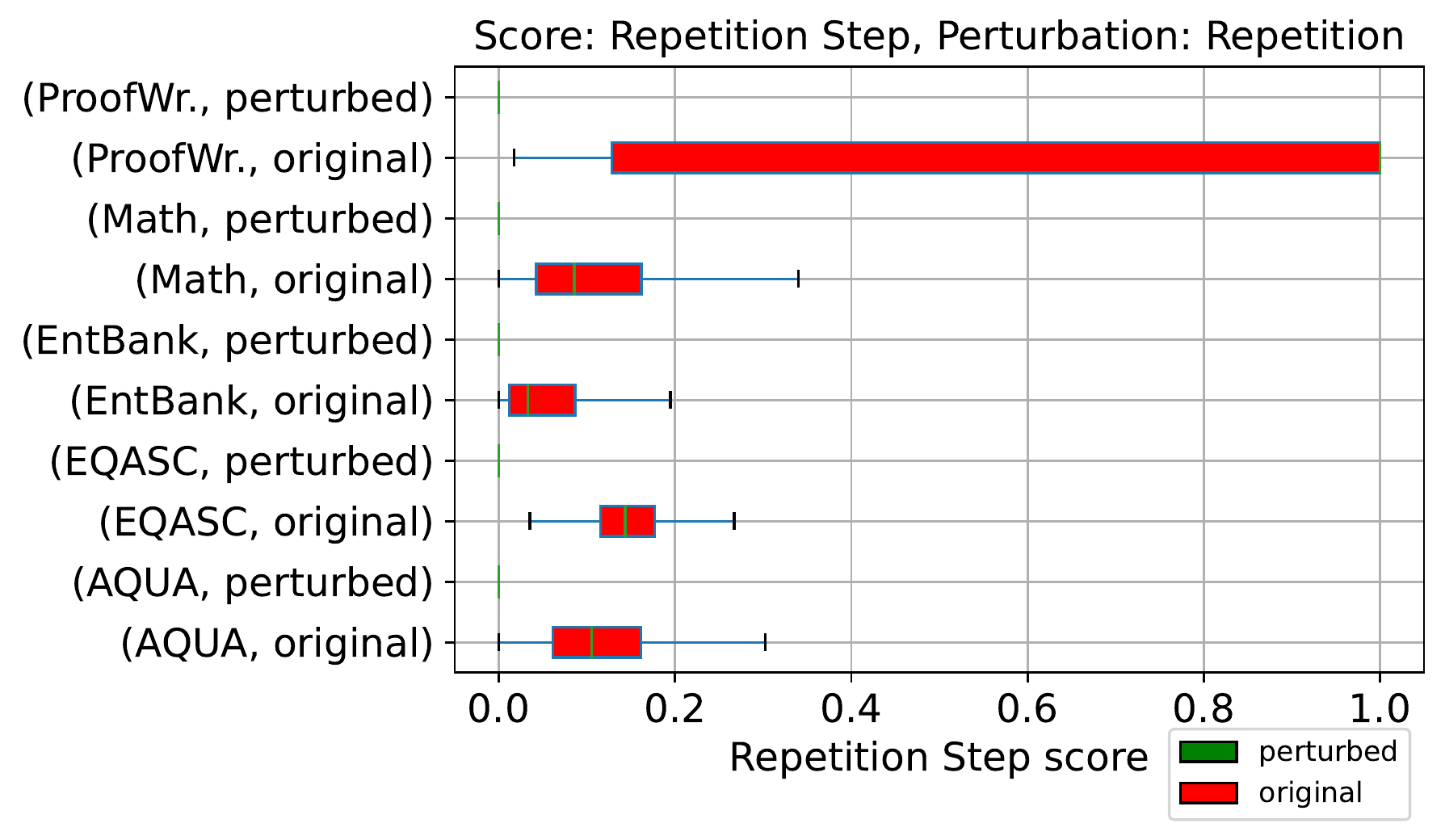}}
    \mbox{\includegraphics[height=.2\textheight]{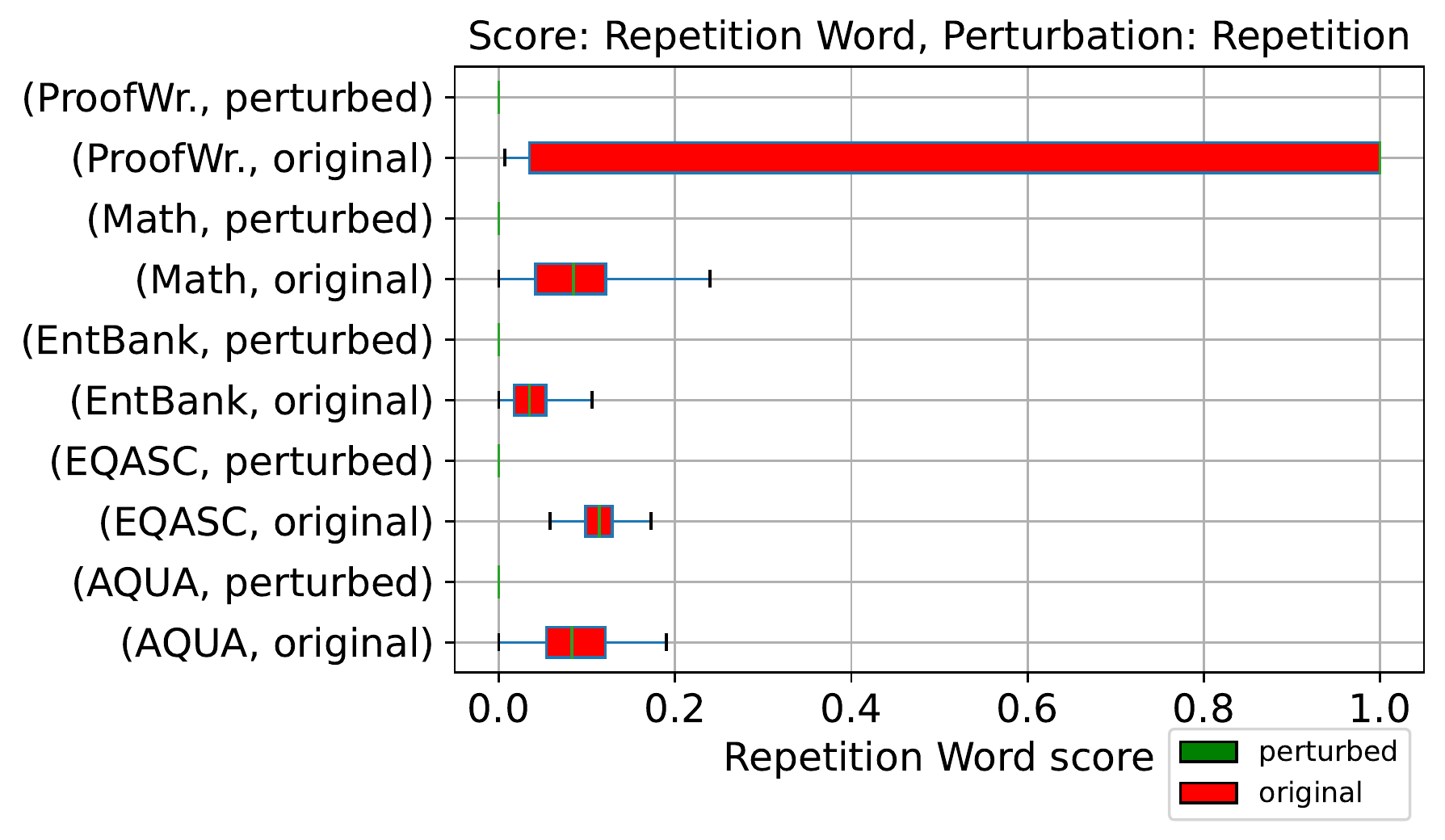}}
    \hspace{-8px}
    \caption{Box-and-whisker plots of interquartile ranges of scores, for \textit{Repetition} perturbations and \textit{Repetition-*} scores. While all perturbed subsets have $0$ or near $0$ scores, all datasets except EQASC have some chains that were also scored as low despite the absence of duplicates.}
    \label{fig:repetition-whiskers}
\end{figure}

Highest correlations among reference-free scores belong to the \textit{Repetition-*} scores, that exhibit perfect correlation on EQASC dataset (Tables~\ref{tab:corr-synth-unsup-perturb-eqasc}-\ref{tab:corr-synth-unsup-perturb-aqua}). For other datasets, non-perfect correlations can be attributed to the small number of false-negatives, i.e. they give low \textit{Repetition-*} scores for chains with non-duplicated but similar steps, while all chains with duplicates got almost $0$ scores (Fig.~\ref{fig:repetition-whiskers}). In EQASC explanations are created from a set of facts that are not directly related to each other, but are intended to give an answer when combined together. Among all datasets considered, these steps are most dissimilar, and thus can be separated with similarity-based scores.

\newpage

\clearpage
\subsection{Experiments with Human Judgement Datasets}
\label{app:human-experiments}

In this section, we present Somers' $D$ correlation of all metrics on all Human Judged datasets in separate tables. Specifically, Table~\ref{tab:corr-human-unsup} summarizes meta-evaluations for \ourmodel~metrics in comparison to baselines on \textbf{all} human judged datasets. Fine-grained evaluations are presented in Table~\ref{tab:corr-human-unsup-DROP} for DROP, 
Table~\ref{tab:corr-human-unsup-GSM8K},~\ref{tab:corr-human-sup-GSM8K} for GSM8K,
Table~\ref{tab:corr-human-unsup-ESNLI},~\ref{tab:corr-human-sup-ESNLI} for ESNLI,
Table~\ref{tab:corr-human-unsup-COSMOS} for CosmosQA, and
Table~\ref{tab:corr-human-unsup-SEMEVAL} for SemEVAL.
Human evaluation perspectives used in evaluations are described in App. Table~\ref{tab:human-eval-perspectives}.

\begin{figure}[h!]      
    \mbox{\includegraphics[height=.4\textheight]{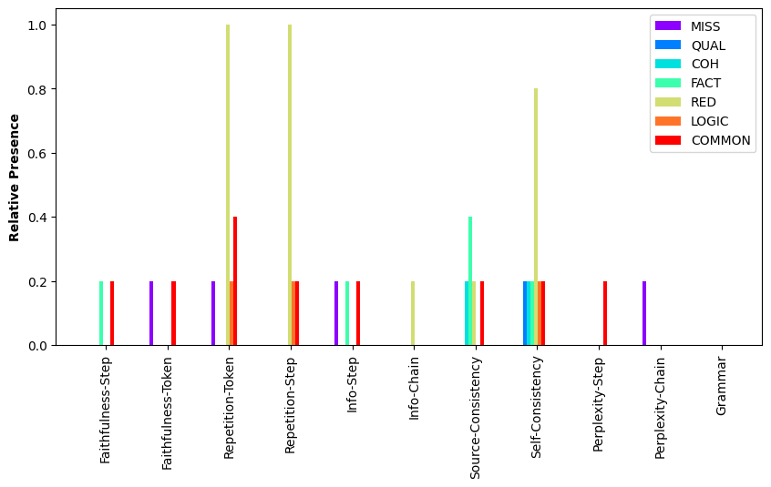}}
    \caption{Relative presence of the strong score-error correlation, measured as the number of datasets where for each score and evaluation perspective pair Somers' $D$ correlation value is in the $90^{th}$ percentile, normalized by the total number of datasets where this type of perturbation occurs. Statistics collected over \ourmodel~reference-free scores with finetuned SimCSE embeddings, and evaluation perspectives where at least 10 errors are present in a dataset.}
    \label{fig:score-ha-error}
\end{figure}

Looking at how errors are captured by \ourmodel~reference-free scores (Fig.~\ref{fig:score-ha-error}), we observe strongest correlations between \textit{Redundancy} error and \textit{Repetition-*}, \textit{Self-Consistency} scores. \textit{Repetition} error is not present in this analysis as it has at most 3 occurrences per dataset. Out of the all considered scores, \textit{Self-Consistency} is able to cover 6 out of 7 evaluation perspectives, except \textit{Missing Step}. 

\input{tables/table-human.tex}
\input{tables/table-DROP.tex}
\input{tables/table-GSM8K.tex}
\input{tables/table-ESNLI.tex}
\input{tables/table-COSMOS.tex}
\input{tables/table-SEMEVAL.tex}
\input{tables/table-GSM8K-sup.tex}
\input{tables/table-ESNLI-sup.tex}

\vspace{0.2cm}
We further look at specific human annotated examples where our ~\ourmodel~gives highest and lowest scores to understand strength and weaknesses of the proposed approach. Results are summarized in Table~\ref{tab:examples}. Similar analysis for diagnostic datasets is summarized in Table~\ref{tab:synthetic-examples}.

\input{tables/table-appx-examples}
\input{tables/table-appx-examples-synthetic}
\end{document}